\documentclass[lettersize,journal]{IEEEtran}
\usepackage{amsmath,amsfonts}
\usepackage[ruled,linesnumbered]{algorithm2e}
\usepackage{changes}
\usepackage{array}
\usepackage{textcomp}
\usepackage{stfloats}
\usepackage{url}
\usepackage{color}
\usepackage{verbatim}
\usepackage{graphicx}
\usepackage{cite}
\usepackage{amssymb}
\usepackage{listings}
\usepackage{enumerate}
\usepackage{multirow}
\usepackage[switch,mathlines]{lineno}
\usepackage{caption}
\usepackage[pagebackref=False,breaklinks=true,letterpaper=true,colorlinks,linkcolor=blue,anchorcolor=blue,citecolor=blue,urlcolor=blue,bookmarks=false]{hyperref}
\usepackage{doi}
\usepackage{lipsum}
\usepackage{pdfpages}

\usepackage{float}
\usepackage[caption=false,font=footnotesize,labelfont=rm,textfont=rm]{subfig}
\usepackage{booktabs}
\usepackage{color}   


\newcommand{\ie}{{i}.{e}., }

\newcommand{\etal}{\textit{~et~al}. }

\newcommand\tableref[1]{Table~\ref{#1}}
\newcommand\myeqref[1]{Eq.~\eqref{#1}}

\begin{document}

\title{A Pairwise Comparison Relation-assisted Multi-objective Evolutionary Neural Architecture Search Method with Multi-population Mechanism}

\author{
    Yu Xue,~\IEEEmembership{Senior Member,~IEEE},
    Pengcheng Jiang,~\IEEEmembership{~Graduate~Student~Member,~IEEE},
    Chenchen Zhu, \\
    MengChu Zhou,~\IEEEmembership{Fellow,~IEEE},
    Mohamed Wahib,
    and Moncef Gabbouj,~\IEEEmembership{Fellow,~IEEE}

\thanks{
	This work was supported by the National Natural Science Foundation of China (NO.62376127, NO.61876089, NO.61876185, NO.61902281), the Natural Science Foundation of Jiangsu Province (BK20141005) and the Natural Science Foundation of the Jiangsu Higher Education Institutions of China (14KJB520025). \textit{(Corresponding author: Yu Xue.)}
	
	Yu Xue, Pengcheng Jiang and Chenchen Zhu are with the School of Software, Nanjing University of Information Science and Technology, Jiangsu, China. E-mails: xueyu@nuist.edu.cn; pcjiang@nuist.edu.cn; 202212490283@nuist.edu.cn.

    MengChu Zhou is with the Department of Electrical and Computer Engineering, New Jersey Institute of Technology, Newark, NJ 07102 USA. E-mail: mengchu@ieee.org.
    
    Mohamed Wahib is with RIKEN Center for Computational Science (R-CCS), Japan. E-mail: mohamed.attia@riken.jp.

    Moncef Gabbouj is with the Department of Computing Sciences, Tampere University, Tampere, Finland. E-mail: moncef.gabbouj@tuni.fi.

}}

\markboth{Journal of \LaTeX\ Class Files,~Vol.~14, No.~8, August~2021}%
{Shell \MakeLowercase{\textit{et al.}}: A Sample Article Using IEEEtran.cls for IEEE Journals}

\maketitle


\begin{abstract}
Neural architecture search (NAS) has emerged as a powerful paradigm that enables researchers to automatically explore vast search spaces and discover efficient neural networks. However, NAS suffers from a critical bottleneck, \ie the evaluation of numerous architectures during the search process demands substantial computing resources and time. In order to improve the efficiency of NAS, a series of methods have been proposed to reduce the evaluation time of neural architectures. However, they are not efficient enough and still only focus on the accuracy of architectures. Beyond classification accuracy, real-world applications increasingly demand more efficient and compact network architectures that balance multiple performance criteria. To address these challenges, we propose the SMEMNAS, a pairwise comparison relation-assisted multi-objective evolutionary algorithm based on a multi-population mechanism.
In the SMEMNAS, a surrogate model is constructed based on pairwise comparison relations to predict the accuracy ranking of architectures, rather than the absolute accuracy.
Moreover, two populations cooperate with each other in the search process, \ie a main population that guides the evolutionary process, while a vice population that enhances search diversity.
Our method aims to discover high-performance models that simultaneously optimize multiple objectives.
We conduct comprehensive experiments on CIFAR-10, CIFAR-100 and ImageNet datasets to validate the effectiveness of our approach.
With only a single GPU searching for 0.17 days, competitive architectures can be found by SMEMNAS which achieves 78.91\% accuracy with the MAdds of 570M on the ImageNet.
This work makes a significant advancement in the field of NAS.
\end{abstract}

\begin{IEEEkeywords}
Evolutionary neural architecture search, surrogate model, multi-objective optimization.
\end{IEEEkeywords}


\section{Introduction}
\IEEEPARstart{C}{onvolutional} neural networks (CNNs) have achieved great success in various machine learning tasks~\cite{bi_network_2024, tan_dynamic_2022}. The performance of neural networks is determined by structures and hyperparameters. Bayesian optimization methods like SMAC3~\cite{lindauer_smac3_2022} are effective hyperparameter optimization methods. The structures of networks are also critical to performance. However, in order to achieve promising performance, traditional CNNs are usually designed manually by human experts with extensive domain knowledge and experience. Not every interested user has such expertise, and even for experts, designing CNNs is also a time-consuming and error-prone process. Therefore, in order to facilitate and automate the design of deep convolutional neural networks, Zoph\etal first proposed the concept of neural architecture search (NAS)~\cite{zoph_neural_2017} and developed a well-known neural architecture search method for CNNs called NASNet~\cite{zoph_learning_2018}, which can obtain neural architectures with highly competitive performance compared to hand-crafted neural architectures. In recent years, researchers have developed many NAS methods, which have attracted increasing attention from both industry and academia in a variety of learning tasks~\cite{ding_bnas-v2_2022, ding_stacked_2023}, such as object detection~\cite{ghiasi_nas-fpn_2019}, semantic segmentation~\cite{nekrasov_fast_2019}, and natural language processing~\cite{magalhaes_creating_2023}.

In addition to the accuracy of neural networks, real-world applications also require NAS methods to find computationally efficient network architectures, e.g., low power consumption in mobile applications and low latency in autonomous driving. Generally speaking, the classification accuracy usually continuously increases with the complexity of the network architectures (i.e., the number of layers, the number of channels, etc.)~\cite{tan_efficientnet_2019}. This implies that maximizing the accuracy and minimizing the network complexity are two competing and conflicting objectives, thus requiring multi-objective optimization for NAS. Existing NAS algorithms can be classified into three categories including reinforcement learning (RL) based~\cite{zoph_learning_2018,pham_efficient_2018}, evolutionary algorithm (EA) based~\cite{sun_completely_2020,lu_multiobjective_2021} and gradient-based methods~\cite{luo_neural_2018,liu_darts_2018}. Despite recent advances in RL-based and gradient-based NAS methods, they are still not easily applicable to multi-objective NAS. In gradient-based methods, the search space is continuous and relies on gradient descent to optimize the neural architectures, while other objectives, such as complexity, can not be optimized by the loss function. So, multiple objectives are not easy to be optimized and the diversity of architectures could be missed in gradient-based methods. RL-based methods are more costly and time-consuming than the other two methods. Evolutionary algorithm is a search method based on evolutionary principles that searches for an optimal solution through the evolution and selection~\cite{sun_completely_2020}. When dealing with multi-objective problems in NAS field, evolutionary algorithms are more adaptable and flexible.

Regarding multi-objective NAS, researchers have proposed many methods for finding high-performance architectures that satisfy multiple metrics simultaneously~\cite{zhang_gpu_2023}. For example, Tan\etal designed a multi-objective neural architecture search approach that optimizes both accuracy and real-world latency on mobile devices~\cite{tan_mnasnet_2019}. Tan and Le designed a new mixed depthwise convolution with accuracy and computation as the optimization metrics~\cite{tan_mixconv_2019}. But these methods have lower efficiency in optimizing architectures. In addition to the problem of search efficiency, there is a lack of improvement in the search strategy. 
During the search process, there is often a lack of diversity in the generated architecture, leading to premature convergence to suboptimal solutions. This phenomenon can be attributed to several factors, including  the selection operation favouring one particular objective to retain individuals and inadequate evolutionary operators. These constraints often hinder the exploration of the entire Pareto front, thereby limiting the quality of the obtained architecture~\cite{tian_multistage_2021}.
For example, NSGANetV1~\cite{lu_multiobjective_2021} tends to search for architectures within a small range around a certain floating-point operations counts (FLOPs), which may lead to the convergence of the local optimal solution. Therefore, how to maintain the diversity of the population without sacrificing performance is a challenging problem\cite{ming_two-stage_2022, yang_local-diversity_2023}.
Lu\etal proposed an efficient method for searching neural network architectures based on the Once-for-All supernet, which allows for rapid screening~\cite{lu_nsganetv2_2020}. However, this method lacks further processing of the population, which prevents it from effectively exploring both objectives during the search process, thereby making it difficult to maintain good population diversity.
There exist a number of diversity preservation approaches in existing multi-objective evolutionary algorithms, like multi-population strategies, which have not been well applied in NAS. Additionally, existing multi-population strategies often use static population selection strategies for the next generation. We consider that different search stages should focus on different search preferences, \ie early strategies for exploitation and late strategies for exploration. Therefore, we improve the selection operation and propose a multi-population mechanism to generate more diverse architectures.

Nevertheless, regardless of the search space and search strategy used, one bottleneck in this field is the need to evaluate a large number of network architectures during the search process, which requires huge computational resources and time~\cite{liu_survey_2022}. Many researchers have proposed lots of methods to improve the efficiency of NAS algorithms, including weight sharing~\cite{pham_efficient_2018,liu_darts_2018}, population memory~\cite{sun_completely_2020}, early stopping mechanism~\cite{sun_particle_2019}, and so on. But most of these methods still have to explicitly or implicitly select numerous architectures and then perform training to evaluate them. Recently, some researchers have established surrogate models to virtually evaluate network architectures, which can execute the environment selection with the predicted accuracy, significantly reducing the evaluation time~\cite{baker_accelerating_2018}. However, one of the key challenges in obtaining reliable surrogate models is that obtaining training samples (pairs of architecture and accuracy) is computationally expensive in some extend. In existing methods, sample utilization of surrogate models is not efficient. For example, Peng\etal trained and evaluated a large number of architectures to be as training samples for the surrogate model in PRE-NAS~\cite{peng_pre-nas_2022}. In NSGANetV2, Lu\etal used a self-adaptive ensemble model to fit the accuracy of the architecture samples with the original encoding of the individuals~\cite{lu_nsganetv2_2020}. But, these methods achieve training samples without special data processing and the quantity is very limited. Moreover, we find that surrogate models are not necessary to produce reliable accuracy estimates (accuracy in the absolute sense), as long as the predicted results are consistent with the true performance ranking of architectures. In the evolutionary algorithm, the selection operation needs to be performed according to the fitness values of the individuals, and only the individuals with high fitness are selected for subsequent evolution. Therefore, it is more important to obtain the ranking of the candidate architectures than to predict the true accuracy.

To address the above problems, we propose an efficient algorithm, called SMEMNAS, which is a surrogate-assisted multi-objective evolutionary algorithm with a multi-population mechanism for NAS in this work. 
The main framework of the proposed algorithm adopts NSGANetV2. The training accuracy of the network and the number of multiply-add operations (MAdds) are selected as two optimization objectives. We propose corresponding measures to address the issues of insufficient population diversity and inadequate predictive capability of the surrogate model.
Firstly, we establish an efficient surrogate model based on pairwise comparison relations to obtain the accuracy ranking of candidate architectures. Then we conduct practical training on the selected top-ranked architectures. Furthermore, to simultaneously optimize the accuracy and network complexity, we design a multi-objective evolutionary algorithm with a multi-population mechanism and a comparison relations-based surrogate model, improving both the efficiency of the search process and the diversity of the resulting architectures. Our approach aims to provide a set of high-performance architectures that take into account multiple optimization objectives.
This work intends to make the following contributions to advance the field of NAS:
\begin{enumerate}[1)]
    \item A more efficient surrogate model based on pairwise comparison relations is proposed. Different from existing surrogate models that focus on absolute accuracy, ours focuses more on the relative rankings among network architectures. It is further able to shorten the time consumption of the search process compared to previous surrogate models.
    \item A multi-objective evolutionary algorithm based on a multi-population mechanism is proposed, in which a main population guides the evolution and a vice population assists the evolution. The mechanism can greatly prevent the algorithm from falling into local optima and speed up the convergence of the algorithm.
    \item We validate the effectiveness of SMEMNAS on the image classification task with standard datasets, CIFAR-10, CIFAR-100 and ImageNet. The proposed method reduces search time to 0.17 GPU days on the ImageNet. The search architecture accuracy outperforms existing methods at the same amount of computation (MAdds).
\end{enumerate}

The remainder of this paper is organized as follows: Section \ref{sec:RelatedWork} presents the related works and background. Section \ref{sec:proposedalgorithm} describes our proposed method in detail. We present the experimental design and results to verify the effectiveness and efficiency of our method with respect to its peers in Section \ref{sec:experiment}. Finally, the conclusions and future works are drawn in Section \ref{sec:conclusion}.

\section{Related Work}\label{sec:RelatedWork}
 In recent years, deep neural networks have been widely deployed in different applications and computing environments. Researches and explorations for NAS are particularly attractive. In the following, Section \ref{sec:multi-objectiveNAS} and Section \ref{sec:surrogatemodels} present studies related to multi-objective NAS and surrogate models for NAS. The introduction and usage of the supernet are presented in Section \ref{sec:supernet}. 

\subsection{Multi-objective NAS}\label{sec:multi-objectiveNAS}
Evolutionary algorithms are widely utilized to tackle multi-objective optimization problems due to their effectiveness in exploring large and complex solution spaces. By simulating the biological evolution, evolutionary algorithms are more flexible to handle multi-objective problems~\cite{chen_diversity_2019}. With the growing demand for efficient models in deep learning, the idea of multi-objective optimization has been introduced to neural architecture search. In the early researches of NAS, most studies only focus on the accuracy of neural networks. This often leads to that the searched networks suffer from excessive computation time and model size. However, the accuracy is rarely the only metric to be considered in real-world applications. It is often necessary to consider other additional and conflicting goals specific to the deployment scenario in the real world, such as model size, model computation, inference latency, and so on. Thus, some researchers have tried to take these secondary objectives into account while searching for architectures with high accuracy. For example, Lu\etal considered both classification accuracy and model computation, and then use the non-dominated sorting genetic algorithm II (NSGA-II) as the multi-objective optimization method~\cite{lu_multiobjective_2021,lu_nsga-net_2019}. Besides, Xue\etal proposed a multi-objective evolutionary algorithm with a probability stack for NAS, which considers accuracy and time consumption~\cite{xue_neural_2023}. Wang\etal adopted a multi-objective particle swarm optimization algorithm to simultaneously optimize classification accuracy and FLOPs~\cite{wang_evolving_2019}. Du\etal further refined the multi-objective optimization process in NAS with the reference point-based environment selection~\cite{tong_neural_2022}. Most of these existing approaches rely on the evaluation of architectures, and are not efficient enough and still time-consuming~\cite{xie_architecture_2024, lv_benchmarking_2023}. Most studies do not improve the multi-objective evolutionary algorithms (MOEAs), which tends to just use the existing multi-objective evolutionary algorithms such as NSGA-II~\cite{lu_nsganetv2_2020}. There exists the phenomenon of small model traps during the search process~\cite{yang_cars_2020}. Specifically, during the early search, smaller models with fewer parameters but higher test accuracy tend to dominate larger models that are less accurate but have the potential for higher accuracy~\cite{ying_nas-bench-101_2019}. Moreover, unsuitable evolution operators or conflicting objective characteristics may also lead to low diversity of the population and the imbalance between diversity and convergence, which may cause the search results to fall into the local optima. The multi-population mechanism is an effective strategy for addressing multi-objective optimization problems~\cite{xu_co-evolutionary_2024}. By maintaining multiple populations, this technique can explore different regions in the search space, solve different optimization tasks simultaneously and so on. Recently, this mechanism has also been adopted in multi-objective NAS~\cite{liao_emt-nas_2023,zou_multiple_2024}. However, existing methods often use multiple populations for different datasets or tasks. For example, Song\etal~\cite{song_multi-population_2024} leverage the classification strengths of various population architectures in their preferred categories to enhance classification accuracy. These multi-population NAS methods do not focus on the issue of diversity. Our research goal is to achieve a balance between accuracy and efficiency in a single task, further enhancing search efficiency and result diversity.

\subsection{Surrogate Models}\label{sec:surrogatemodels}
The main computational bottleneck in NAS is the performance evaluation of architectures. In recent years, many surrogate-based methods have been proposed to accelerate the evaluation process in ENAS~\cite{song_surrogate-assisted_2023}. Liu\etal from Google used the long short-term memory (LSTM) as a surrogate model to replace the training of deep neural networks~\cite{liu_progressive_2018}. Afterwards, Luo\etal proposed a neural network architecture optimization method, NAO, by using MLP as a surrogate model which works better than PNAS~\cite{luo_neural_2018}. Sun\etal used an offline surrogate model based on random forest (RF) to predict the performance directly for the encoded architectures~\cite{sun_surrogate-assisted_2020}. Guo\etal exploited a ranking loss function to learn a predictor model that predicts the relative score of architectures~\cite{guo_generalized_2022}. They can directly score and rank the performance of architectures to assist the evaluation process in NAS. However, most existing surrogate models, including those designed as accuracy predictors, are simplistic and fail to fully utilize training samples. For instance, accuracy predictors, which directly estimate the accuracy of architectures, often suffer from low generalization due to limited training data. This inaccuracy can disrupt the predicted rankings (i.e., rank disorder), thereby compromising the reliability of environment selection during the search process. Recently, Wang\etal proposed a neural architecture search method based on particle swarm optimization, which uses support vector machine (SVM) as a surrogate model to conduct the preliminary exploration on the individual comparison relationship~\cite{wang_surrogate-assisted_2022}. Despite the impressive progress, they search in a specific space constructed by themselves rather than in a general search space. This makes their method difficult to compare with related methods, and limits its generalizability.

\subsection{SuperNet}\label{sec:supernet}
Supernet is a powerful technique within NAS. The supernet is a hyper-set of all the candidate networks in the search space, which means that each candidate network can be regarded as being formed by some modules or paths from the supernet. With such technique, there are two mainly used methods to efficiently search. One of the methods uses progressive training during the search.
DARTS~\cite{liu_darts_2018} updates the architecture parameters by gradients and it can also be reflected in the supernet. Yang\etal\cite{yang_cars_2020} used NSGA-III to search for architectures and only updates the parameters in the subnet during training of every step. This method can efficiently search for potential candidates without a pre-trained supernet. Another kind of methods is the one-shot method, which relies on a pre-trained supernet.
Guo\etal~\cite{guo2020single} proposed the single path one-shot neural architecture search, SPOS. They trained a supernet, and then sampled and evaluated each sub-network. The emergence of FairNAS~\cite{chu_fairnas_2021} and GreedyNAS~\cite{you2020greedynas} further expanded the search methods based on the supernets. 
Once-For-All~\cite{cai_once_2020} is another widely used method for one-shot NAS. It supports variations in four factors: depth, width, convolutional kernel size, and image resolution. The Once-For-All supernet is firstly trained by the progressive shrinking algorithm, in which the largest network is fine-tuned to support sub-networks and they are added to the sampling space by sharing weights. The weights inherited from the trained supernet are used as a warm-up for the gradient descent algorithm during our architecture search.

\begin{figure*}
    \centering
    \includegraphics[width=\textwidth]{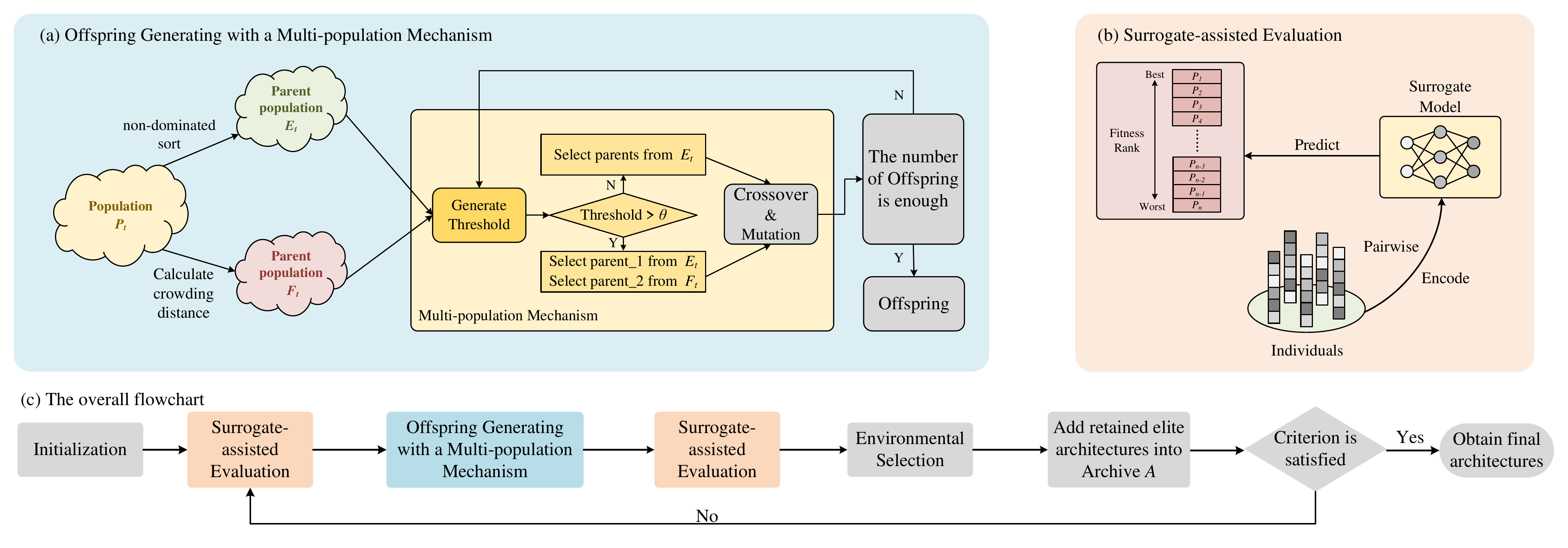}
    \captionsetup{labelfont={color=black}}
    \caption{Overall framework of the multi-objective algorithm based on a multi-population mechanism and the surrogate model assisting the search process (SMEMNAS). (a) Offspring are generated with a multi-population mechanism. (b) The illustration of the evaluation assisted by the surrogate model. (c) The flowchart of the proposed algorithm. A set of elite architectures are retained in each generation and are added into the archive $\mathcal{A}$, which contains architectures for evolution.
    }
    \label{fig:flowchat}
\end{figure*}
\begin{algorithm}
\SetAlgoLined
\caption{SMEMNAS}\label{algo:Framework}
\KwIn{Search space $S$, number of initial samples $N$, SuperNet $S_W$, number of iterations $T$, complexity objective $f_c$, number of selected candidates $K$, candidates population $R_t$.}
$i \leftarrow 0$ \tcp{Initialize a sampling individual counter.} 
$t \leftarrow 0$ \tcp{Initialize a iteration counter.}
$\mathcal{A} \leftarrow \phi$ \tcp{Initialize an empty archive to store evaluated architectures.} \label{algo:Framework:3}
\While{$i < N$}{\label{algo:Framework:4}
\tcp{Sample $N$ architectures Randomly.}
$\alpha \leftarrow $ Sample($S$)

$\omega_\alpha \leftarrow S_W$($\alpha$) \tcp{weights inherited from the trained supernet.}
$acc \leftarrow $ SGD($\alpha$, $\omega_\alpha$)

$\mathcal{A} \leftarrow \mathcal{A}$ $\cup$ ($\alpha$, acc)

$i \leftarrow i + 1$
}\label{algo:Framework:10}
\While{$t < T$}{\label{algo:Framework:11}
$predictor \leftarrow $ Construct surrogate model from $\mathcal{A}$\tcp{Online update the $predictor$.}\label{algo:Framework:12}

$Solutions \leftarrow $ MP-MOEA($\mathcal{A}$, $predictor$, $f_c$)\label{algo:Framework:13}


$R_t \leftarrow $ Select $K$ elite architectures from $Solutions$

\For{$\alpha $ in $R_t$}{\label{algo:Framework:15}
\tcp{Add candidates to $\mathcal{A}$.}
$\omega_\alpha \leftarrow S_W$($\alpha$)

$acc \leftarrow $ SGD($\alpha$, $\omega_\alpha$)

$\mathcal{A} \leftarrow \mathcal{A}$ $\cup$ ($\alpha$, acc)
}\label{algo:Framework:19}
$t \leftarrow t + 1$
}
\Return architectures chosen by NonDominatedSort($\mathcal{A}$) \label{algo:Framework:22}
\end{algorithm}

\section{The Proposed Method of Evolutionary Multi-objective Neural Architecture Search}\label{sec:proposedalgorithm}
This section presents the details of our proposed pairwise comparison relation-assisted multi-objective evolutionary algorithm for NAS. We firstly present the framework of the proposed algorithm in Subsection \ref{sec:framework}, and the details of the proposed search space and encoding, surrogate model, and multi-objective algorithm with a multi-population mechanism are presented in Subsections \ref{sec:encoding} to \ref{sec:MP-MOEA}.

\subsection{Overall Framework}\label{sec:framework}
An overview of the proposed algorithm is illustrated in Fig. \ref{fig:flowchat}. As shown in Fig. \ref{fig:flowchat}, our algorithm generally follows the basic process of genetic algorithm. Compared to existing NAS methods, the proposed method has two key innovations: one is that a surrogate model is constructed to assist the evaluation. Another is that a multi-population mechanism is added into the process of generating new individuals.

Algorithm \ref{algo:Framework} shows the pseudo code of the proposed algorithm. Firstly, in order to construct an efficient surrogate model, some evaluated architectures are needed to construct the training dataset. So before the search process starts, an empty archive $\mathcal{A}$ for storing training samples is initialized (line \ref{algo:Framework:3}). $N$ individuals are randomly sampled from the search space and then are decoded for training. As mentioned above, we employ the weight sharing during the training. The weights inherited from the trained supernet are used as a warm-up for the stochastic gradient descent (SGD) algorithm to improve the search efficiency (lines \ref{algo:Framework:4}-\ref{algo:Framework:10}). Afterwards, there is a main search loop of in the algorithm (line \ref{algo:Framework:11}). The surrogate model is constructed using the archive $\mathcal{A}$ (line \ref{algo:Framework:12}). A detailed description of the surrogate model is given later in subsection \ref{sec:surrogate}. Then we define a set to store individuals to be evaluated. Next, we use the proposed multi-objective evolutionary algorithm with multi-population mechanism (MP-MOEA for short) combine with our surrogate model to optimize both accuracy (Acc) and MAdds, i.e., maximize classification accuracy and minimize model complexity (line \ref{algo:Framework:13}). This process is described in detail later in subsection \ref{sec:MP-MOEA}. To improve the sample efficiency of our search, we train the surrogate model using an online learning approach. The retained individuals of each generation are decoded and trained, and these retained offspring are used as new training samples to update the surrogate model (lines \ref{algo:Framework:15}-\ref{algo:Framework:19}). The above steps are repeated until the conditions for the end of the procedure are satisfied. Finally, after the multi-objective evolutionary search, the optimal architectures are selected from the pool of architectures based on the non-dominated sorting (line \ref{algo:Framework:22}).

\begin{figure*}
    \centering
    \includegraphics[width=0.75\textwidth]{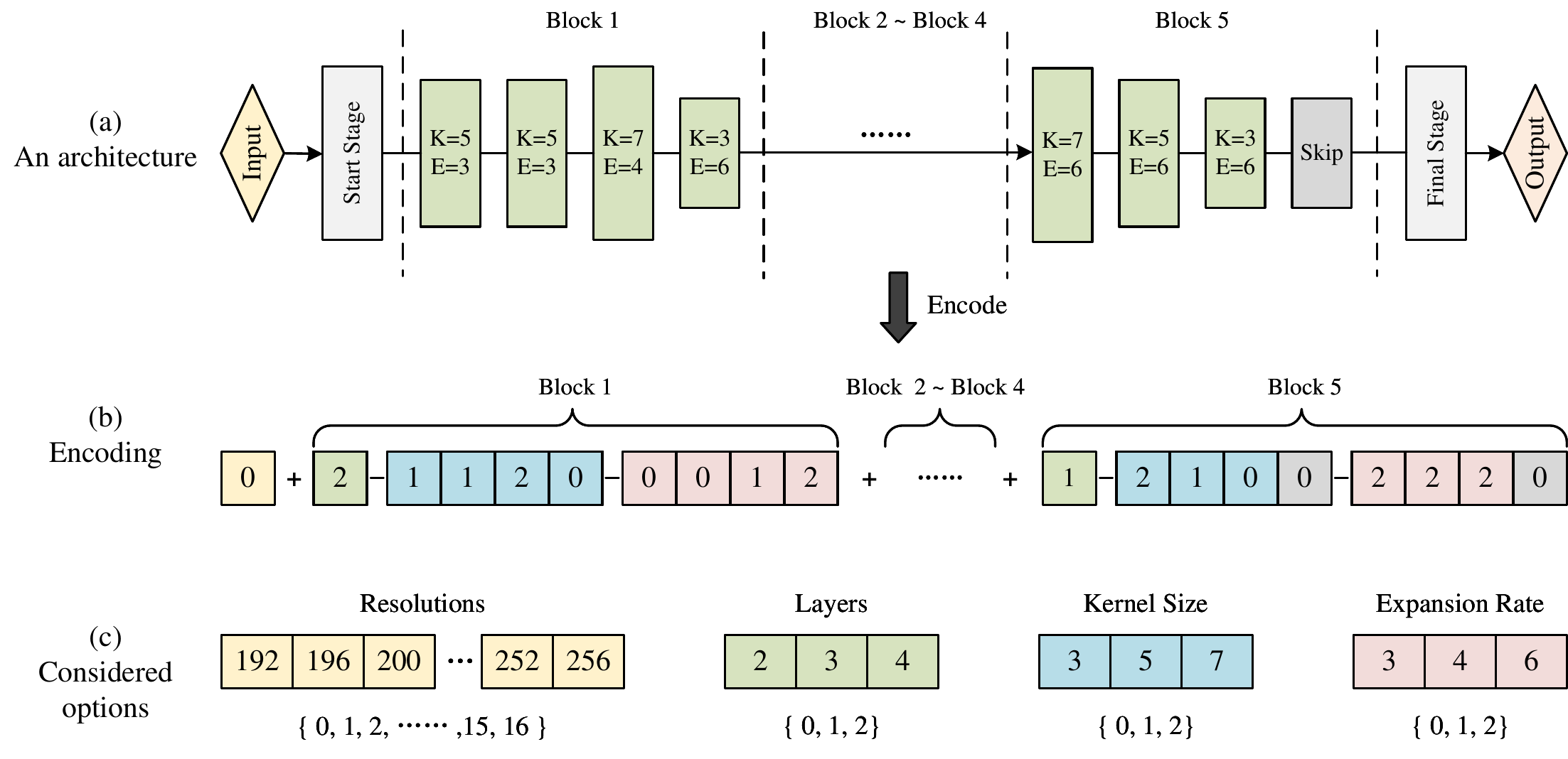}
    \caption{\textbf{A candidate architecture encoding example.} The encoding is divided into five parts by blocks. The parameters we search include image resolution, the number of layers in each block, the expansion rate and the kernel size in each layer.}
    \label{fig:encoding}
\end{figure*} 

\begin{figure}[h]
    \centering
    \includegraphics[width=0.35\textwidth]{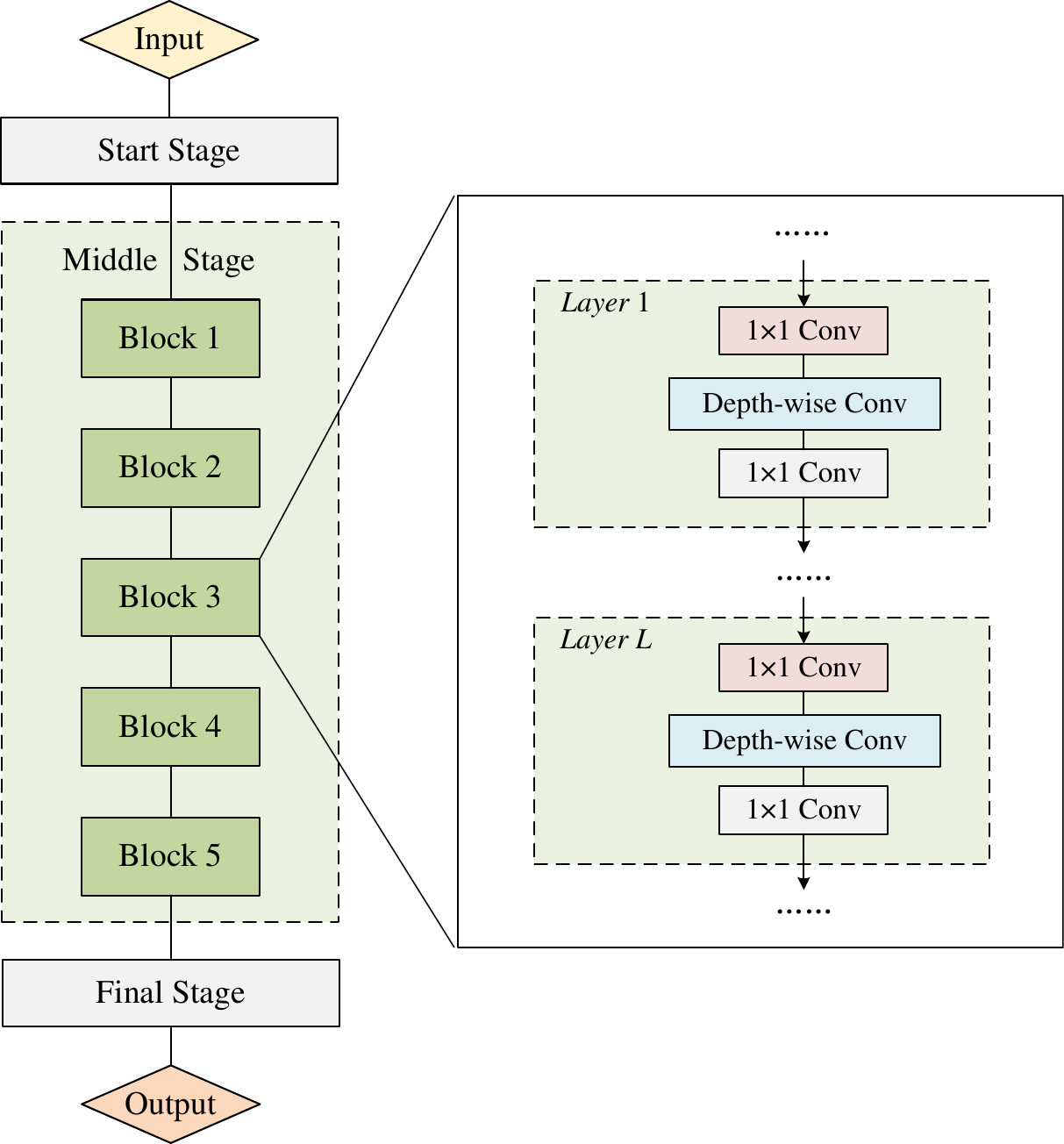}
    \caption{\textbf{Search space:} The left part shows a complete network stacked by five blocks. The right part represents a Block and its internal structure, which consists of multiple layers.}
    \label{fig:searchspace}
\end{figure}

\subsection{Search Space and Encoding}\label{sec:encoding}
The search process of NAS starts with constructing a search space that can contain most CNNs for image tasks. In this paper, the backbone structure is based on MobileNetV3~\cite{howard_searching_2019}. The structure consists of three stages. The start stage extracts features, and the final stage outputs categories. These two parts do not need to be searched. The middle stage needs to be searched and consists of five sequentially connected MBConvBlocks~\cite{cai_once_2020} that progressively decrease the feature map size and increase the number of channels. Each block is composed of a series of layers, and every layer adopts the inverted bottleneck structure~\cite{sandler_mobilenetv2_2018}: first, a $1\times1$ convolution, which is used to convert the input channel to the expansion channel; second, a depth-wise convolution, which is the expansion channel and contains the parameter stride; at last, a $1\times1$ convolution, which is used to convert from the expansion channel to the output channel. 
The number of layers of each convolution block (depth) is selected from $\left\{ 2, 3, 4 \right\}$; for each layer, we search the expansion rate of the first $1\times1$ convolution and the kernel size of the depth-wise separable convolution. The expansion rate is selected from $\left\{ 3, 4, 6 \right\}$, and the kernel size is selected from $\left\{ 3, 5, 7 \right\}$. Moreover, we also allow the CNNs to obtain multiple input image size (resolution), ranging from 192 to 256 with a stride of 4. Therefore, there are approximately $10^{20}$ different architectures in the search space. For the encoding strategy, we use a fixed 46-bit integer string. If the encoding length of the architecture with fewer layers is less than 46 bits, the fixed length is achieved by padding with zeros. Fig. \ref{fig:encoding} shows the encoding strategy, and Fig. \ref{fig:searchspace} shows the search space of this paper. As shown in these figures, the architecture is composed of five blocks. The encoding of this architecture is composed of image resolution and other parts representing five blocks. For each block, its encoding consists of the number of layers, the expansion rate, and the kernel size of each layer. In addition, the values in the encoding are all the index of considered options. As depicted in Fig. \ref{fig:encoding}, the first bit ``0" denotes the index of the first candidate resolution which means the resolution is 192. The remaining 45 bits represent five blocks, each of which is represented by 9 bits. e.g., the Block 1 can be encoded as ``211200012". The first bit ``2" denotes the index of  candidate layers which means the Block 1 has four layers. Then, the next four bits ``1120" denote the kernel size of each layer. Specifically, the kernel size in each layer in order is $\left\{5, 5, 7, 3 \right\}$. Similarly, the rest four bits denote that the expansion rate in each layer in order is $\left\{3, 3, 4, 6 \right\}$. In Block 5, the gray bit ``0" represents no layer and the encoding is padded with zeros, which is not from considered options.

\subsection{Surrogate Model}\label{sec:surrogate}
Unlike most of the existing surrogate-assisted methods used to predict the classification accuracy of architectures, we use the surrogate model to obtain the performance ranking of candidate architectures. In the evolutionary algorithm, the selection operation needs to be carried out according to the fitness values of individuals, and individuals with high fitness will be selected for the subsequent evolution. Therefore, it is more important to obtain the ranking of the candidates than to predict the true accuracy. We propose a surrogate model based on pairwise comparison relations, which is constructed on the basis of a classification model. Its inputs are vectors of the encoding combinations of pairwise architectures, and its outputs are labels indicating which one is the better solution. In this paper, we try to study the comparison relation between two individuals and transform the comparison relation learning problem into a binary classification problem. Binary classification models can be trained better with fewer samples. For example, given $n$ labeled architectures, we generate training samples by comparing all unique pairs of architectures, and $n(n-1)/2$ training samples can be obtained. Given a small number of samples, the surrogate model with better performance can be obtained through the extended training samples.

\begin{figure}[t]
    \centering
    \includegraphics[width=0.5\textwidth]{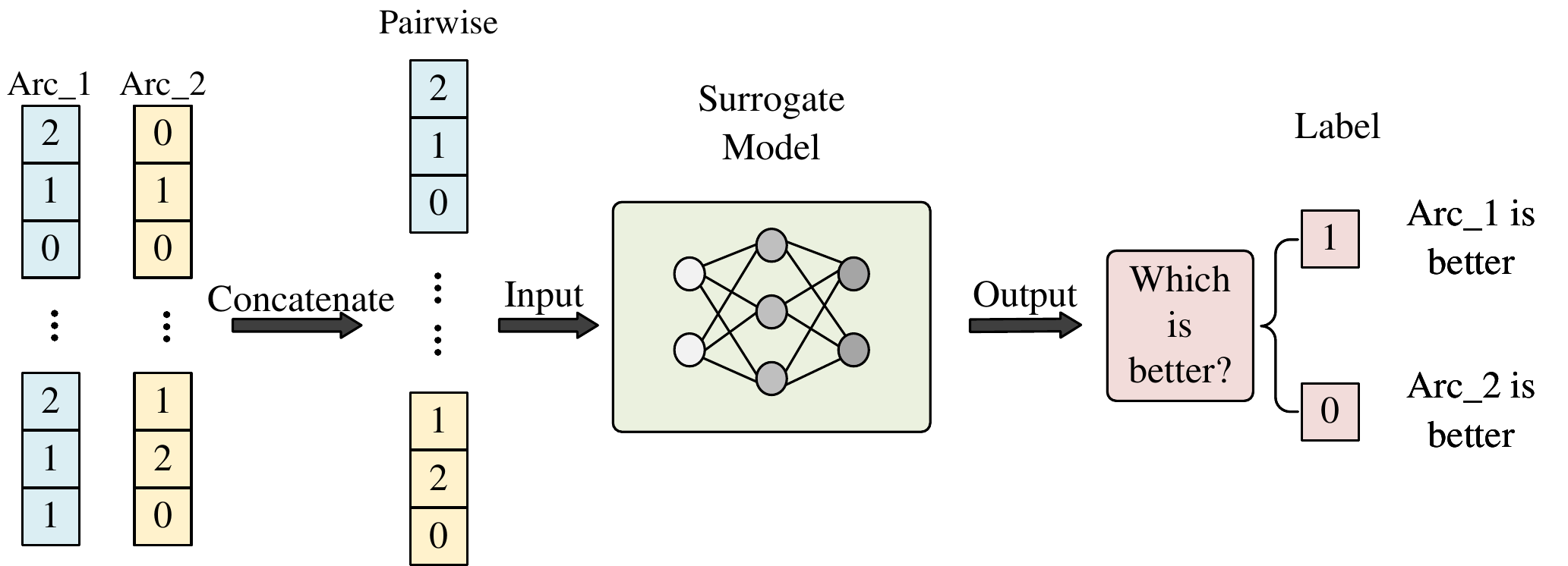}
    \captionsetup{labelfont={color=black}}
    \caption{\textbf{An illustration of the surrogate model based on pairwise comparison relation:} the input contains the concatenation of two architectures, and the output indicates which architecture is better.}
    \label{fig:surrogate}
\end{figure}

In this paper, SVM is used as the surrogate model. In order to train the surrogate model, we should firstly construct the training dataset. The construction method is the same as that proposed in EffPNet~\cite{wang_surrogate-assisted_2022}. The specific process is as follows: firstly, the data used to train needs to be built from the archive $\mathcal{A}$. The $n$ individuals in the archive $\mathcal{A}$ are matched in pairs sequentially. In a pair of individuals, if the fitness value of the first individual is better than the other, the class label will be 1. Otherwise, the class label will be 0. Then $n(n-1)/2$ training samples can be obtained to feed into the SVM for training. The above data processing transforms the direct prediction of architecture performance into a binary classification task comparing the performance of a pair of architectures. We take a pair of encodings of two architectures as the input of the surrogate model, and the output is label ``1" or ``0" as described above. To make it more intuitive, an example is provided in Fig. \ref{fig:surrogate} to present the concatenation of the encodings of two architectures, as well as the output. Label ``1" indicates that architecture $Arc\_1$ has better performance, while label ``0" indicates that architecture $Arc\_2$ has better performance.

Through the above pairwise relations prediction, we can obtain the performance ranking of all architectures. The details are as follows: assuming that there are $n$ architectures, we set a score for each architecture corresponding to its performance. Firstly, we match the first architecture with the other $n-1$ architectures and use the trained surrogate model to predict the relationship of these pairs of architectures. For each pair of architectures, if the first architecture outperforms the second one, the score of the first architecture is added by 1, otherwise the score of the second architecture is added by 1. Then, the second architecture is paired one by one with the remaining architectures after removing the first architecture, and the above process is repeated until the last architecture is left. Finally, each architecture is compared $n-1$ times and the score ranges from 0 to $n-1$. The number of comparisons in total is $n(n-1)/2$. The larger the value of the score, the more times the corresponding architecture wins in the overall comparisons, which implies that the architecture has a superior performance. The results from pairwise relations prediction will be used in the non-dominated sorting for the environment selection during the subsequent evolutionary process.

\subsection{Multi-objective Evolutionary Algorithm Based on Multi-population Mechanism}\label{sec:MP-MOEA}
In multi-objective evolutionary algorithms, diversity needs to be considered as well as convergence. During the practical evolution process, it is easy to fall into the local optimum without exploring the region of the optimal solution. To overcome this problem, we propose a multi-objective evolutionary algorithm based on the multi-population mechanism, see Algorithm \ref{algo:MP-MOEA} for details. The multi-population mechanism is mainly used in the selection operation to select parents for the next generation during the evolution process. Firstly, two sub-populations are constructed from the initial population $P_t$ (line \ref{algo:MP-MOEA:1}). The main population is $E_t$, and the vice population is $F_t$. Specifically, the initial population $P_t$ is non-dominated sorted according to the predicted architectural rankings obtained from the above surrogate model and the calculated MAdds. After that, the non-dominated solutions of the first level are taken as the main population $E_t$ (lines \ref{algo:MP-MOEA:2}-\ref{algo:MP-MOEA:5}). Then, the individuals in the population $E_t$ are removed from the population $P_t$, and the crowding distances of the remaining solutions are calculated. $K$ individuals with high crowding ranking are selected to form the vice population $F_t$ (line \ref{algo:MP-MOEA:6}). We set a threshold based on the number of generations, and when this threshold is less than $\theta$, Both parent $p_1$ and $p_2$ come from the main population $E_t$ (line \ref{algo:MP-MOEA:14}). If this threshold is greater than $\theta$, parent $p_1$ comes from the primary population and parent $p_2$ comes from the secondary population (line \ref{algo:MP-MOEA:16}). A number of offspring architectures are generated until the number of offspring is greater than or equal to $m$ in the current generation. $m$ is a number used to limit the number of offspring. The threshold is a value related to the evolution generation, which is generated by the following equation:

\begin{align}\label{eq:Threshold}
 Threshold=& \left\{
\begin{array}{lc}
random(\delta, 0.7) &  g < \frac{1}{4}G \\
random(0, \delta)  &  \frac{1}{4}G \leq g \leq \frac{3}{4}G \\
random(\delta, 1) &  g > \frac{3}{4}G \\
\end{array} \right.     \\
\delta =& \left\{
\begin{array}{lc}\label{eq:Threshold2}
random(0, 0.7) &  g < \frac{1}{4}G \\
random(0, 1)   &  g \geq \frac{1}{4}G \\
\end{array} \right.
\end{align}

\noindent where $Threshold$ and $\delta$ are randomly generated in every generation, $g$ denotes the serial number of the current generation, and $G$ is the total number of generations.

\begin{algorithm}[t]
\SetAlgoLined
\caption{Proposed MP-MOEA}\label{algo:MP-MOEA}
\KwIn{Initial population $P_t$, surrogate model $predictor$, complexity objective $f_c$, number of generations $G$, vice population size $K$.}
$P_t \leftarrow $ architectures in $\mathcal{A}$ \tcp{$\mathcal{A}$ from Algo.\ref{algo:Framework}.} \label{algo:MP-MOEA:1}
Predict the accuracy ranking of architectures in $P_t$ by $predictor$ \tcp{$predictor$ see Algo.\ref{algo:Framework}.}\label{algo:MP-MOEA:2}
Evaluate the complexity $f_c$ of each architecture in $P_t$\\
NonDominatedSort($P_t$) according to the predicted ranking and complexity\\
$E_t \leftarrow $ All non-dominated solutions in $P_t$
\tcp{main population.} \label{algo:MP-MOEA:5}

$F_t \leftarrow $ Select $K$ solutions from the rest dominated solutions in $P_t$ by crowding distance \tcp{vice population.} \label{algo:MP-MOEA:6}

$g \leftarrow 0$\\
$Q_t \leftarrow \phi$ \tcp{offspring population.}
\While{$g < G$}{
\tcp{Multi-population Mechanism.}
$\theta \leftarrow 0.5$\\
$Threshold \leftarrow $ Generated by Eq.(1)\\
\While{the number of architectures in $Q_t < m$}{
\eIf{$Threshold \textgreater \theta$}{
$p_1, p_2 \leftarrow $ Randomly select one solution $p_1$ from $E_t$, another solution $p_2$ from $F_t$\\ \label{algo:MP-MOEA:14}
}
{ 
$p_1, p_2 \leftarrow $ Randomly select two different solutions from $E_t$\\ \label{algo:MP-MOEA:16}
}
$p_1', p_2' \leftarrow $ Crossover($p_1$, $p_2$)\\
$q_1, q_2 \leftarrow $ Mutation($p_1'$), Mutation($p_2'$)\\
$Q_t \leftarrow Q_t \cup q_1 \cup q_2$
}
$P_t \leftarrow P_t \cup Q_t$\\
Predict the accuracy ranking of architectures in $P_t$\\
Evaluate $f_c$ of each architecture in $P_t$\\
NonDominatedSort($P_t$)\\
$E_t \leftarrow $ All non-dominated solutions in $P_t$\\
$F_t \leftarrow $ Update by crowding distance\\
$g \leftarrow g + 1$
}
\Return $P_t$
\end{algorithm}

Integers are used to encode the architectures in this paper. The crossover operator uses two-point crossover. Because the traditional polynomial mutation is based on real numbers, we rewrite it to fit the integer encoding. In the early stage of the search, we should focus on the convergence and diversity. It is vital to keep the whole population evolving in the right direction, so the main population should be the dominant one, and the vice population should play an assisting role in the exploration. In the middle stage of the search, more consideration is given to convergence, and the parents should be the best individuals. So, parents are all selected from the main population $E_t$, which helps to achieve a fast convergence rate. In the later stage, the population evolution tends to be stable, we use the vice population to explore more solution regions and increase the diversity of the population to avoid the algorithm falling into local optima. In the main loop of the algorithm, the main and vice populations will be iteratively updated, and they will collaborate to achieve optimization of the correlation metric with higher accuracy and lower network complexity.

\section{Experiments}\label{sec:experiment}
In this section, we conducted a series of experiments to validate the effectiveness of our proposed method SMEMNAS. Specifically, we firstly show our experimental configurations in Section \ref{sec:configurations}. Next, Section \ref{sec:results} presents and analyses our experimental results on benchmark datasets. Then we evaluate the effectiveness of the multi-population mechanism in Section \ref{sec:resultsMP}. The comparison results with other algorithms in terms of different perspectives are discussed, including accuracy, complexity, search cost, device and year of publication. Finally, Section \ref{sec:surrogate_ablation} shows some ablation experiments for surrogate models and the initial population.

\subsection{Experimental Configurations}\label{sec:configurations}
Regarding the benchmark datasets used in the experiments, we used CIFAR-10, CIFAR-100 and ImageNet datasets consistent with the existing ENAS works. In this study, the process of searching for architectures was performed separately on these three datasets. The performance of the final searched CNN architectures was also evaluated on three datasets. We evaluated the effectiveness of different architectures in terms of both classification accuracy and computational complexity.

For each dataset, we started with 100 randomly sampled architectures from the search space as the initial population. The evolutionary search process was executed for a total of 25 iterations. Except for the initial population, for each subsequent iteration, we retained and evaluated eight architectures from generated offspring. They were used for the training of the surrogate model and for the next iteration. So, in the whole search process, we evaluated 300 architectures in total, which were also used to train the surrogate model. Since evaluating architecture performance is time-consuming, we adopted the weight-sharing technique to accelerate the training process and improve the search efficiency. The Once-For-All~\cite{cai_once_2020} supernet is trained with the progressive shrinking algorithm, in which the largest network is fine-tuned to support sub-networks and they are added to the sampling space by sharing weights. During the search process, the architecture weights were inherited from the trained Once-For-All as initial weights for training. Because Once-For-All is based on the ImageNet dataset, we fine-tuned the weights inherited from the supernet for five epochs when searching on CIFAR-10 and CIFAR-100. All our experiments were performed on a single Nvidia RTX 3090 GPU card. Finally, we selected three promising architectures from the obtained non-dominated solutions and then trained them for 200 epochs by the standard SGD optimizer with momentum, where the initial learning rate, the momentum rate and the batch size were set to 0.01, 0.9 and 128. More detailed information can be found in our publicly available code at \href{https://github.com/pcjiang1998/SMEM-NAS}{https://github.com/pcjiang1998/SMEM-NAS}.

\begin{table}[t]
    \centering
    \caption{Comparison with state-of-the-art image classifiers on the CIFAR-10 dataset. The multi-objectives used for architecture optimization are performance and model complexity. SMEMNAS(w/o): Architectures found by SMEMNAS without the multi-population mechanism. Search cost: GPU days of the searching process.}
    \label{tab:cifar10_result}
    \scalebox{0.7}{
    \begin{tabular}{c|c|c|c|c|c|c|c}
    \hline\hline
        \textbf{Architecture}     & \begin{tabular}[c]{@{}c@{}}\textbf{Top-1}\\\textbf{(\%)}\end{tabular} & \begin{tabular}[c]{@{}c@{}}\textbf{MAdds}\\\textbf{(M)}\end{tabular} & \begin{tabular}[c]{@{}c@{}}\textbf{Params}\\\textbf{(M)}\end{tabular} & \begin{tabular}[c]{@{}c@{}}\textbf{Search}\\\textbf{Cost}\end{tabular} & \textbf{Device}                       & \textbf{Year} & \begin{tabular}[c]{@{}c@{}}\textbf{Search}\\\textbf{Method}\end{tabular}  \\ \hline
        EfficientNet-B0~\cite{tan_efficientnet_2019} & 98.1 & 387 &4.0 & - &- & 2019& manual \\ 
        \hline
        iDARTS~\cite{wang_idarts_2023} & 97.75 & - &3.6 & 0.2 &Tesla V100 & 2023& GD\\
        CDARTS~\cite{yu_cyclic_2023} & 97.52 & - &3.9 & 0.3 &Tesla V100 & 2023& GD \\
        SWD-NAS~\cite{xue_self-adaptive_2024} & 97.49 & 519 &3.17 & \textbf{0.13} &RTX 3090 & 2024& GD\\
        PA-DARTS~\cite{xue_improved_2024} & 97.59 & 578 &3.75 & 0.36 &RTX 3090 & 2024& GD\\
        GENAS~\cite{xue_gradient-guided_2024} & 97.55 & 504 &3.53 & 0.26 &RTX 3090 & 2024& GD\\
        RelativeNAS~\cite{tan_relativenas_2023} & 97.66 & - &3.93 & 0.4 &GTX 1080Ti & 2023& GD\\
        NPENAS-NP~\cite{wei_npenas_2023} & 97.46 & - &3.5 & 1.8 &RTX 2080Ti & 2023& EA \\
        CGP-NAS~\cite{garcia-garcia_continuous_2023} & 96.3 & 636 &4.04 & 11.5 &RTX 1080Ti & 2023& EA\\
        SMCSO~\cite{xue_surrogate_2025} & 97.12 & - &3.46 & 1.32 &RTX 2080Ti & 2025& EA\\
        SPNAS~\cite{jiang_score_2025} & 98.2 & - &6.33 & 1.4 &RTX 3090 & 2025& EA\\
        FairNAS-A~\cite{chu_fairnas_2021} & 98.2 & 391 &- & 12 &Tesla V100 & 2021& EA \\ 
        CARS~\cite{yang_cars_2020} & 97.38 & 728 &3.6 & 0.4 &Tesla V100 & 2020& EA \\ 
        ESENet~\cite{qiu_efficient_2023} & 96.44 & 540 &4.53 & 9&RTX 2080Ti & 2023& EA\\
        NSGANetV2~\cite{lu_nsganetv2_2020}& \textbf{98.4} & 468&- & -&- & 2020& EA \\ 
        M2M-Net~\cite{tan_m2m-net_2024} & 97.56 & - &3.79 & 6 &RTX 3080 & 2024& EA\\
        \hline
        NASNet-A~\cite{zoph_learning_2018} & 97.35 & 608 &3.3 & 1800 &Tesla P100 & 2018& RL \\ 
        BNAS~\cite{ding_bnas_2022} & 97.03 & - &4.7 & 0.19 &- & 2022& RL\\
        DBNAS~\cite{yang_deeply_2025} & 97.58 & - &\textbf{2.4} & 0.9 &RTX 3090 & 2025& RL\\
        \hline
        SMEMNAS-S (w/o) & 97.51 & \textbf{186} &4.7 & 1.25&RTX 3090 & -& EA \\ 
        SMEMNAS-M (w/o) & 97.78 & 313 &5.8 & 1.25 &RTX 3090 & -& EA \\ 
        SMEMNAS-L (w/o) & 97.85 & 755 &6.4 & 1.25 &RTX 3090 & -& EA \\ 
        SMEMNAS-S (ours) & 97.96 & 200 &4.7 & 1.25 &RTX 3090 & -& EA \\ 
        SMEMNAS-M (ours) & 98.03 & 259 &5.7 & 1.25 &RTX 3090 & -& EA \\ 
        SMEMNAS-L (ours)& 98.13 & 305 &5.0 & 1.25 &RTX 3090 & -& EA \\
    \hline\hline
    \end{tabular}
    }
\end{table}

\subsection{Results on Standard Datasets}\label{sec:results}
In this section, we show the results of our method on benchmark datasets to validate its effectiveness. We conducted the search and trained the final architectures separately on three datasets. Then, in order to validate the effectiveness of the proposed method, we compare the non-dominated architectures obtained by SMEMNAS with those obtained by other state-of-the-art NAS methods. The selected peer methods can be broadly divided into three categories: manually designed by human experts, EA-based, and non-EA-based (e.g., Bayesian, RL and GD). In addition, we give the searched results by SMEMNAS without the multi-population mechanism. The results on CIFAR-10 and ImageNet are presented in Tables \ref{tab:cifar10_result} and \ref{tab:imagenet_result}, and the results in CIFAR-100 are presented in the \textbf{supplementary materials} due to the page limitation.

\begin{figure}[t]
    \centering
    \includegraphics[width=0.4\textwidth]{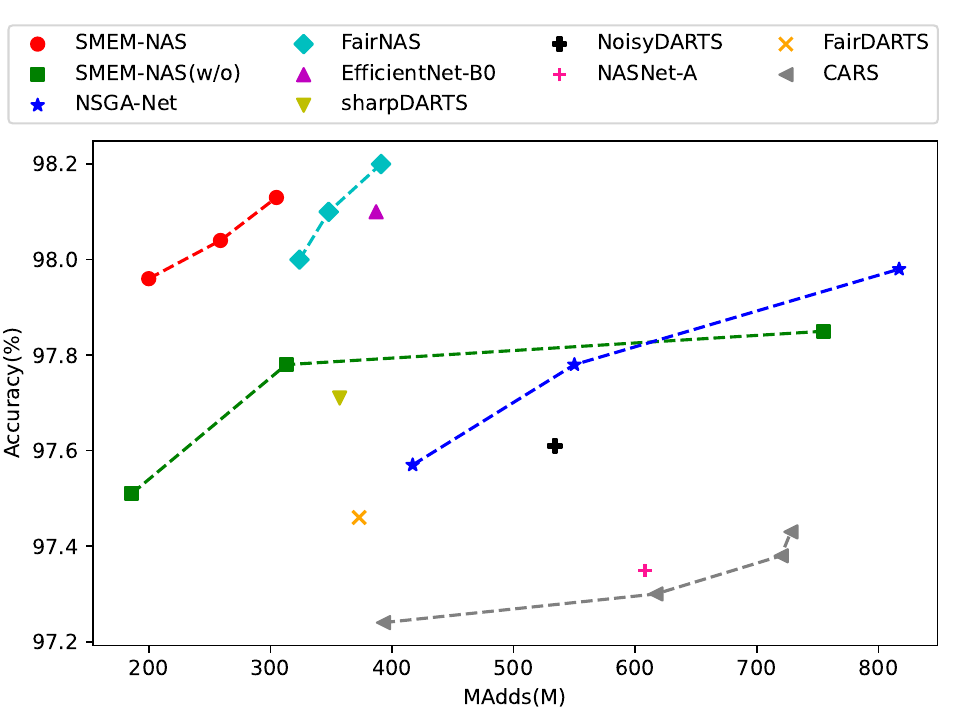}
    \caption{\textbf{Accuracy and number of multi-adds in millions on CIFAR-10.} Models from multi-objective approaches are joined with lines.}
    \label{fig:cifar10}
\end{figure}

\textbf{Results on CIFAR-10:} After completing the search using our method, we selected three promising architectures, and for ease of representation we named them SMEMNAS-S/M/L (sorted by MAdds). Fig. \ref{fig:cifar10} and \tableref{tab:cifar10_result} show the results of SMEMNAS and other methods on CIFAR-10. Fig. \ref{fig:cifar10} visualizes the dominant relationship between the network models in terms of accuracy and model complexity. It is obvious that our models always outperform most other models. \tableref{tab:cifar10_result} shows more details. Firstly, from the perspective of time, our method takes 1.25 GPU days and significantly reduces the search time compared with methods based on EA and RL. The surrogate model we used effectively accelerates the search process. Compared with the GD-based method, although there are some deficiencies in time, the architecture we searched has better performance. Our model achieves the accuracy of 98.13\%, surpassing almost all peer competitors. Among them, our model is slightly lower than FairNAS-A, but has nearly 100M lower MAdds and the search time is much less than FairNAS. In general, SMEMNAS is superior to or consistent with other models in different metrics. Moreover, we also present the results obtained without employing the multipopulation mechanism (denoted as w/o), thereby demonstrating the effectiveness of this mechanism.

\begin{table*}[h]
    \centering
    \caption{Comparison with state-of-the-art image classifiers on the ImageNet dataset. The search cost excludes the supernet training cost.}
    \label{tab:imagenet_result}
    \scalebox{0.85}{
    \begin{tabular}{c|c|c|c|c|c|c|c|c}
    \hline\hline
        \textbf{Architecture} & \textbf{Top-1 Acc (\%)} & \textbf{Top-5 Acc (\%)} & \textbf{MAdds (M)} &\textbf{Params (M)} & \textbf{Search Cost (GPU Days)} & \textbf{Device} & \textbf{Year}& \textbf{Search Method} \\ \hline
        EfficientNet-B0~\cite{tan_efficientnet_2019} & 76.3 & 93.2 & 390 &5.3 & - &- & 2019& manual \\
        
        \hline
        Optuna (so)~\cite{akiba_optuna_2019} & 78.61 & 93.98 & 732 &7.66 & 0.19 &RTX 3090 & 2019& Bayesian\\
        Optuna-M (mo)~\cite{akiba_optuna_2019} & 77.48 & 93.3 & 329 &5.29 & 0.16 &RTX 3090 & 2019& Bayesian\\
        Optuna-L (mo)~\cite{akiba_optuna_2019} & 78.63 & 94.01 & 675 &7.66 & 0.16 &RTX 3090 & 2019& Bayesian\\
        \hline
        PA-DARTS~\cite{xue_improved_2024} & 75.3 & 92.25 & - &5.2 & 0.4 &RTX 3090 & 2024& GD\\
        RelativeNAS~\cite{tan_relativenas_2023} & 75.12 & 92.3 & 563 &5.05 & 0.4 &GTX 1080Ti & 2023& GD\\
        $\beta$-DARTS~\cite{ye_-darts_2022} & 76.1 & 93.0 & 609 &5.5 & 0.4 &- & 2022& GD \\ 
        NAP~\cite{ding_nap_2022} & 75.5 & 92.6 & 574 &4.8 & 4 &RTX 2080Ti & 2022& GD \\ 
        iDARTS~\cite{wang_idarts_2023} & 75.3& 92.3 & 568 &5.1 & 1.9 &Tesla V100 & 2023& GD\\
        SQNAS~\cite{gm_sequential_2024} & 73.8& 91.4 & 540 &- & 0.25&Tesla V100 & 2024& GD\\
        Shapely-NAS~\cite{xiao_shapley-nas_2022} & 76.1 & - & 582 &5.4 & 4.2 &- & 2022& GD \\ 
        RandomNAS~\cite{ma_defying_2025} & 74.2 & - & - &5.9 & 0.5 &-& 2025& GD\\
        \hline
        SPNAS~\cite{jiang_score_2025} & 78.62 & 94.07 & 687 &6.6 & 0.37 &RTX 3090 & 2025& EA\\
        Once-For-All~\cite{cai_once_2020} & 76.9 & 93.2 & 230 &- & 2 &Tesla V100 & 2020& EA \\ 
        FairNAS~\cite{chu_fairnas_2021} & 77.5 & - & 392 &5.9 & 12 &- & 2021& EA \\ 
        
        NSGANetV2~\cite{lu_nsganetv2_2020} & 77.5 & 93.5 & 225 &6.1 & 1 &-& 2020& EA \\
        MixPath~\cite{chu_mixpath_2023} & 77.2 & 93.5 & 378 &5.1 & 10.3 &Tesla V100 & 2023& EA \\
        DPS~\cite{li_extensible_2023} & 75.6 & 92.7 & 578 &5.3 & \textbf{0.06} &RTX A6000 & 2023& EA \\
        CENAS-C~\cite{ma_pareto-wise_2024} & \textbf{79.6} & \textbf{94.4} & 482 &4.98 & 1.98 &RTX 2080Ti & 2024& EA \\
        SLE-NAS-B~\cite{huang_split-level_2023} & 75.7 & 92.5 & 412 &4.5 & 3 &RTX A6000 & 2023& EA\\\hline
        MnasNet~\cite{tan_mnasnet_2019} & 76.7 & 93.3 & 403 &5.2 & 4.5 &64*TPUv2 & 2019& RL \\
        BNAS~\cite{ding_bnas_2022} & 74.3 & 92.2 & - &\textbf{3.9} & 0.19 &- & 2022& RL\\
        DBNAS~\cite{yang_deeply_2025} & 77.6 & 93.5 & 386 &4.9 & 0.9 &RTX 3090 & 2025& RL\\
        \hline
        SMEMNAS-S (ours) & 77.75 & 93.75 & 360 &5.9 & 0.17 &RTX 3090 & -& EA \\
        SMEMNAS-M (ours) & 78.16 & 93.92 & 468 &6.6 & 0.17&RTX 3090 & -& EA \\
        SMEMNAS-L (ours) & 78.91 & 94.23 & 570 &7.0 & 0.17 &RTX 3090& -& EA \\ \hline
        \hline
    \end{tabular}
    }
\end{table*}

\textbf{Results on ImageNet:} Different from the conclusion on the CIFAR datasets, there is no need for pre-training before searching on the ImageNet dataset. \tableref{tab:imagenet_result} shows the results of the proposed method and other comparative methods. SMEMNAS outperforms most other methods and achieves the accuracy of 78.91\% with the MAdds of 570M. In addition, our proposed surrogate model effectively reduces the search time, and only takes 4 hours to complete the search. It is worth mentioning that although the final accuracy achieved by our method is somewhat lower than that of CENAS-C, our search time and resource consumption are significantly lower than CENAS-C. We obtained the final results using only 10\% of the search time with just one GPU. Additionally, we conducted experiments using the Optuna method with the same number of true evaluations on our search space. In \tableref{tab:imagenet_result}, we present the final training results, and it can be observed that even within the same search space, the highest-accuracy architecture obtained by SMEMNAS surpasses the classification accuracy achieved by the Optuna method while reducing MAdds by 100M. For a more detailed comparison between SMEMNAS and Optuna methods, please refer to the \textbf{supplementary materials}.

\begin{figure*}
\centering
\subfloat[Architectures distribution (main 0.7, vice 0.3)]{
\includegraphics[width=0.25\linewidth]{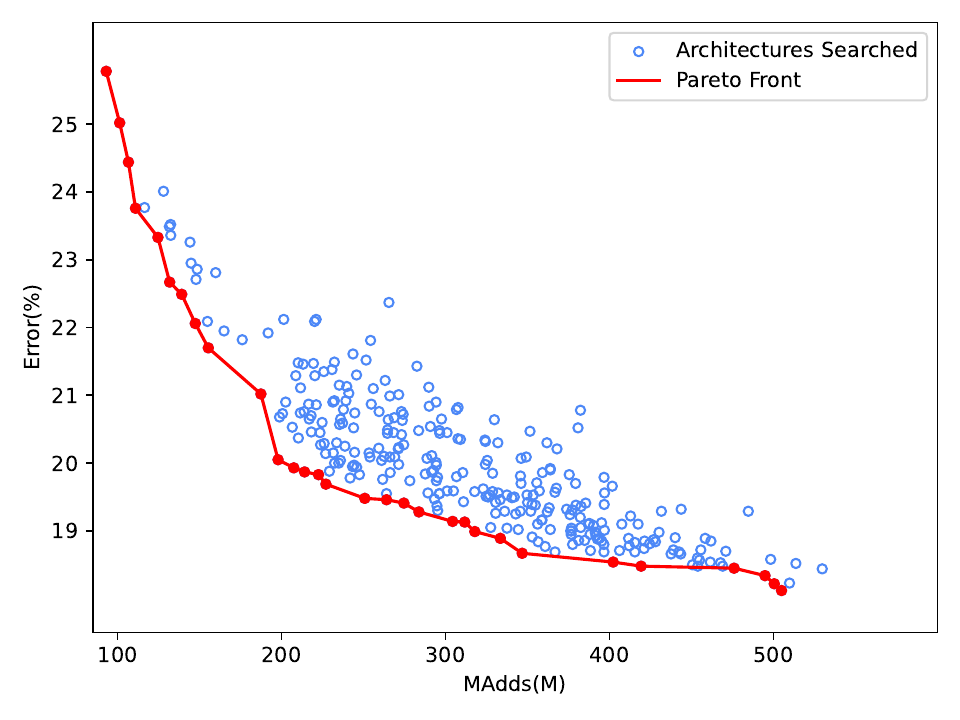}
\label{fig:threshold1}
}
\hspace{4mm}
\subfloat[Architectures distribution (main 0.3, vice 0.7)]{
\includegraphics[width=0.25\linewidth]{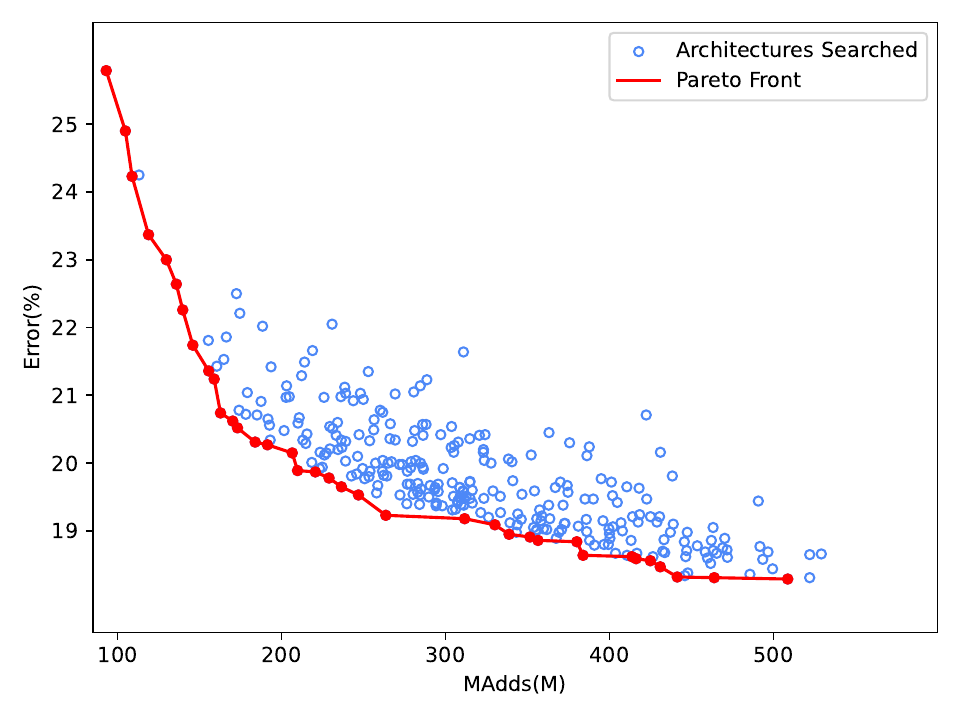}
\label{fig:threshold2}
}
\hspace{4mm}
\subfloat[Architectures distribution (both 0.5)]{
\includegraphics[width=0.25\linewidth]{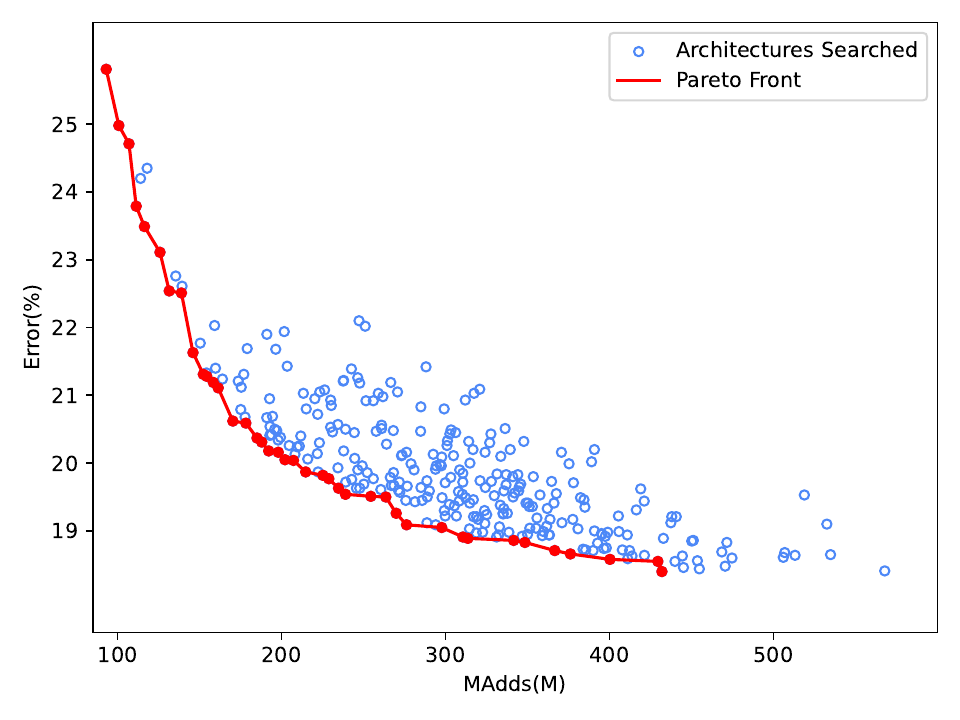}
\label{fig:threshold3}
}
\caption{The comparison of results between different probabilities for using two populations on the ImageNet. Different probability values in the captions of sub-figures indicate the probability that the parents generating the next generation of offspring will come from the two populations. We plot the distribution of architectures in the archive during the search process in three figures. The horizontal and vertical axes are MAdds and error, respectively.}
\label{fig:threshold}
\end{figure*}
\begin{figure}[htbp]
\centering
\subfloat[Architectures distribution Without MP]{
\includegraphics[width=0.45\linewidth]{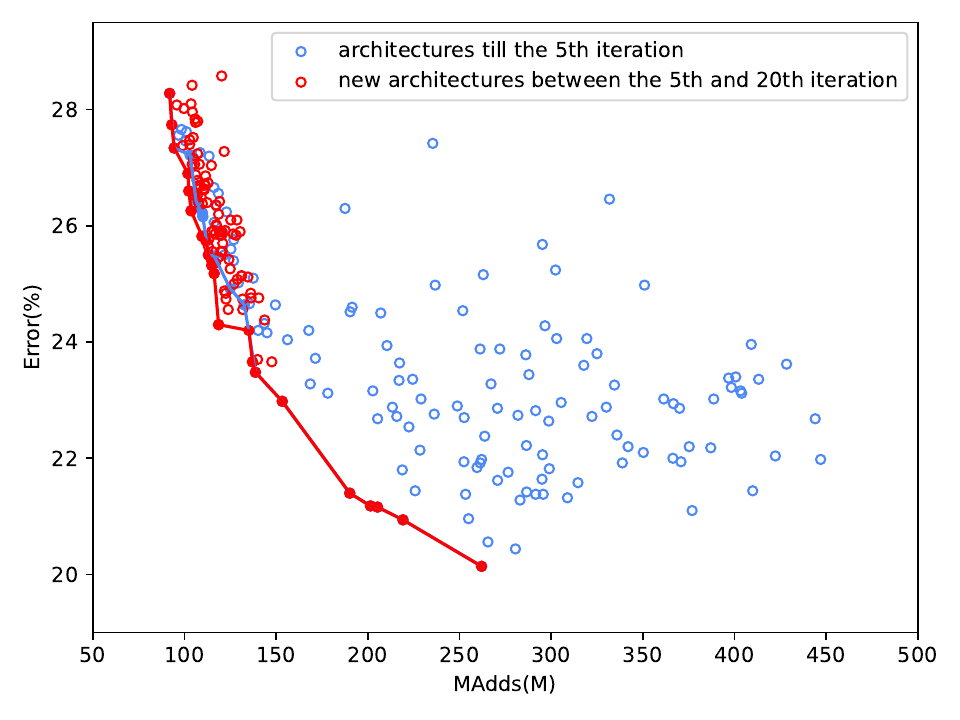}
\label{fig:comparison_wo}
}
\hspace{2mm}
\subfloat[Architectures distribution With MP]{
\includegraphics[width=0.45\linewidth]{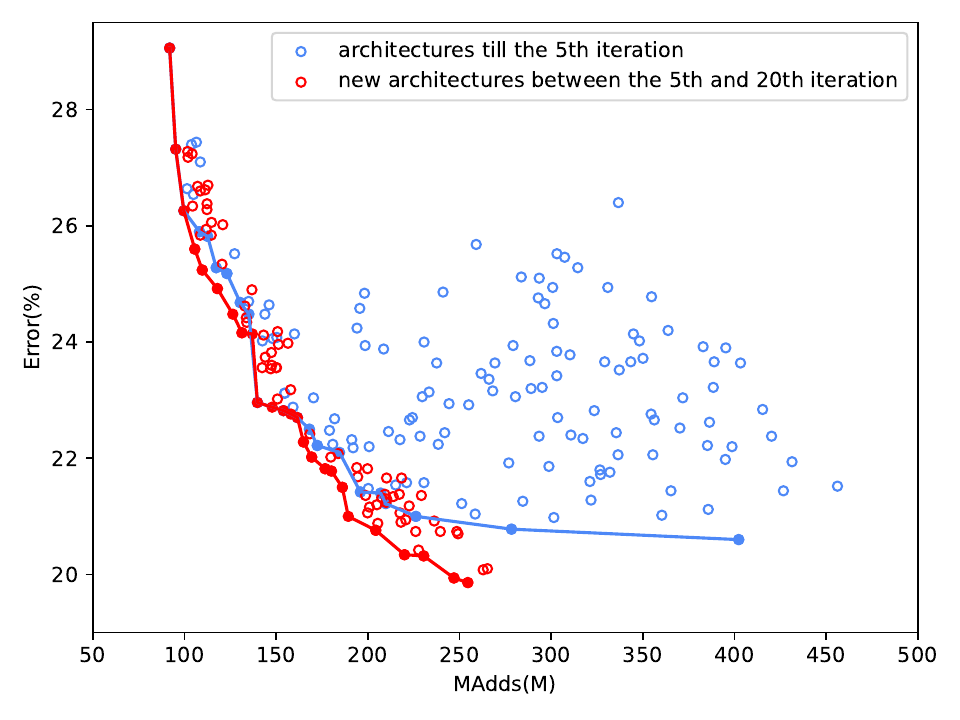}
\label{fig:comparison_MP}
}
\caption{The comparison between SMEMNAS with and without the multi-population mechanism (MP) on CIFAR-100. We plot the distribution of architectures in the archive during the search process in two figures.}
\label{fig:comparison}
\end{figure}

\begin{figure*}[htbp]
\centering
\subfloat{
\includegraphics[width=0.22\linewidth]{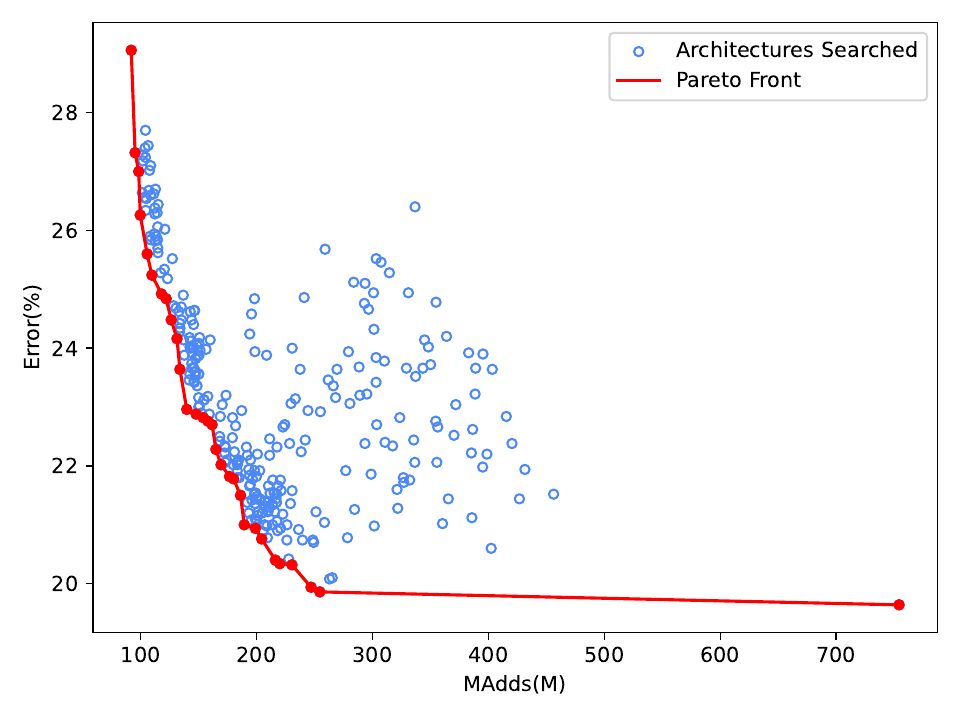}
}
\subfloat{
\includegraphics[width=0.22\linewidth]{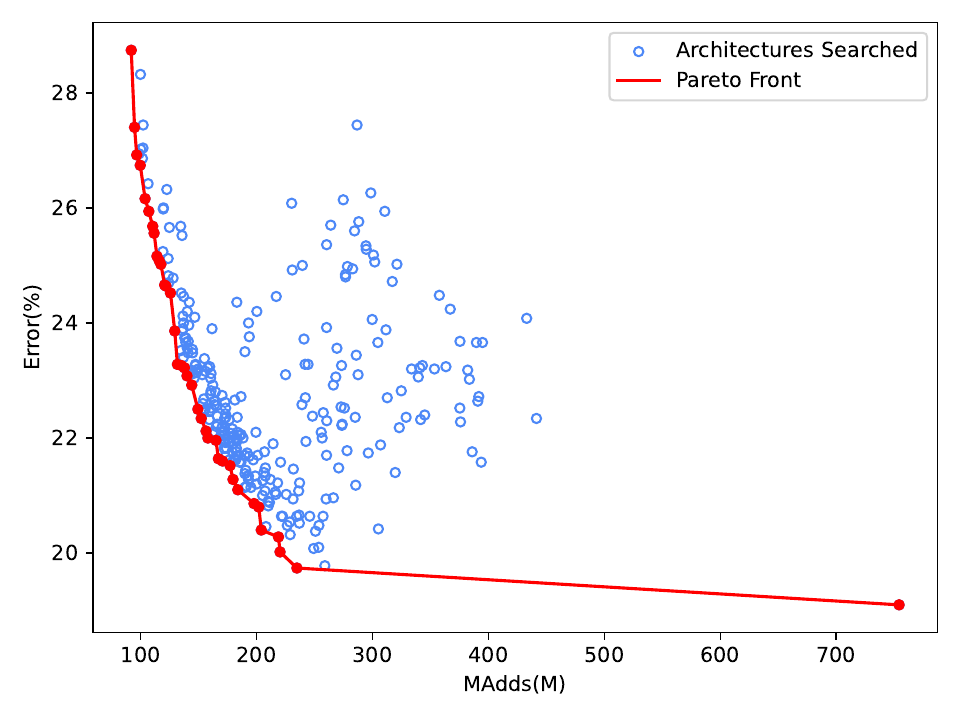}
}
\subfloat{
\includegraphics[width=0.22\linewidth]{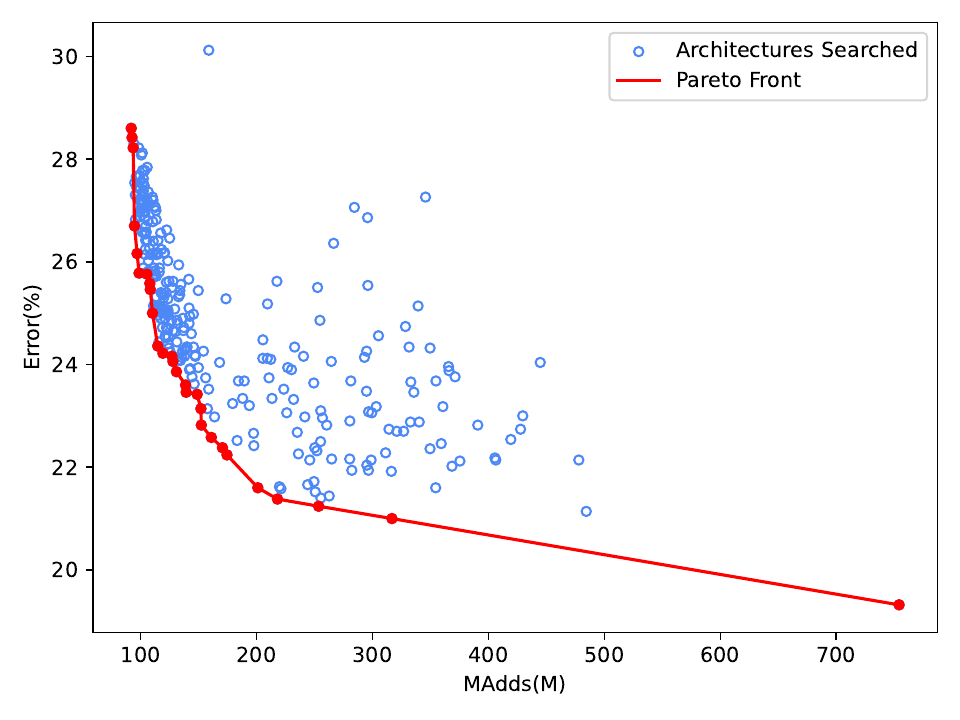}
}
\subfloat{
\includegraphics[width=0.22\linewidth]{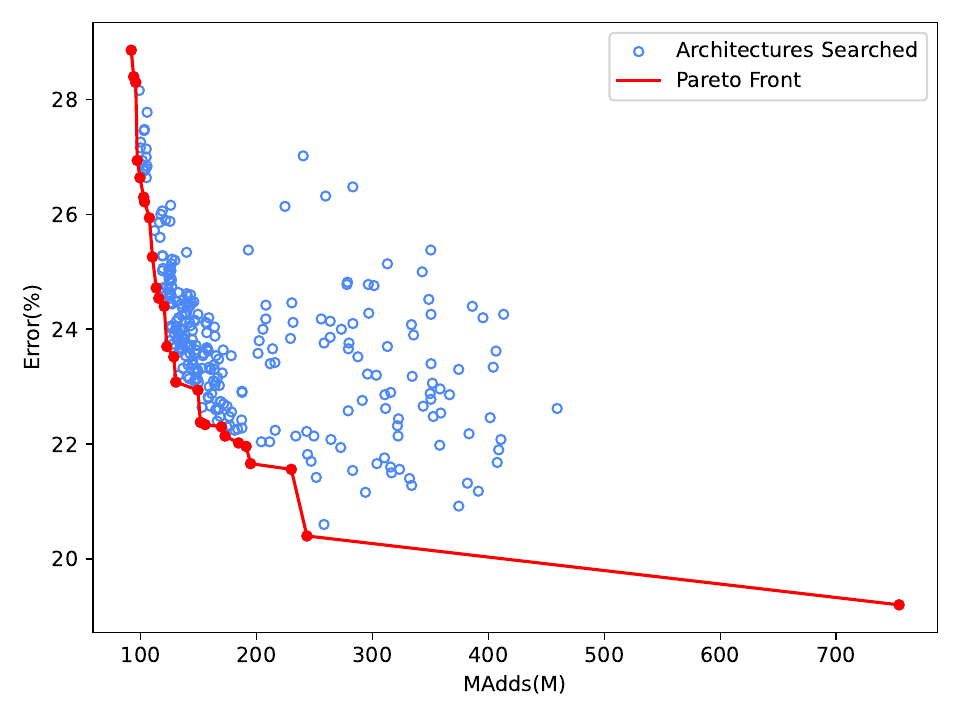}
}

\vspace{-12pt}
\setcounter{subfigure}{0}
\subfloat[SVM ($KTau$ = 0.93)]{
\includegraphics[width=0.22\linewidth]{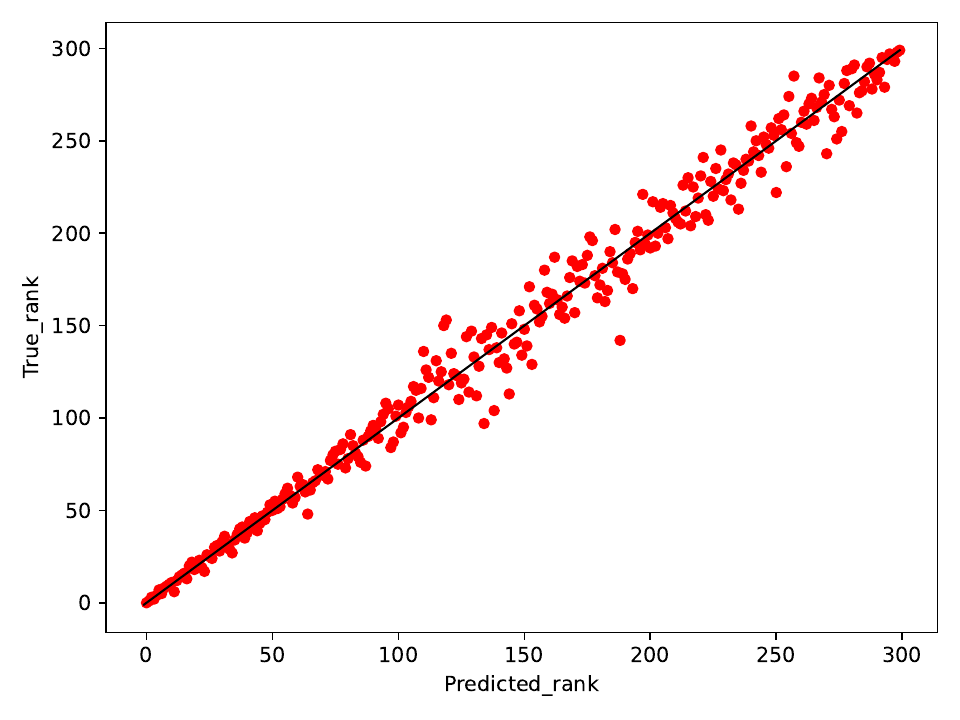}
\label{SVM}
}
\subfloat[MLP ($KTau$ = 0.91)]{
\includegraphics[width=0.22\linewidth]{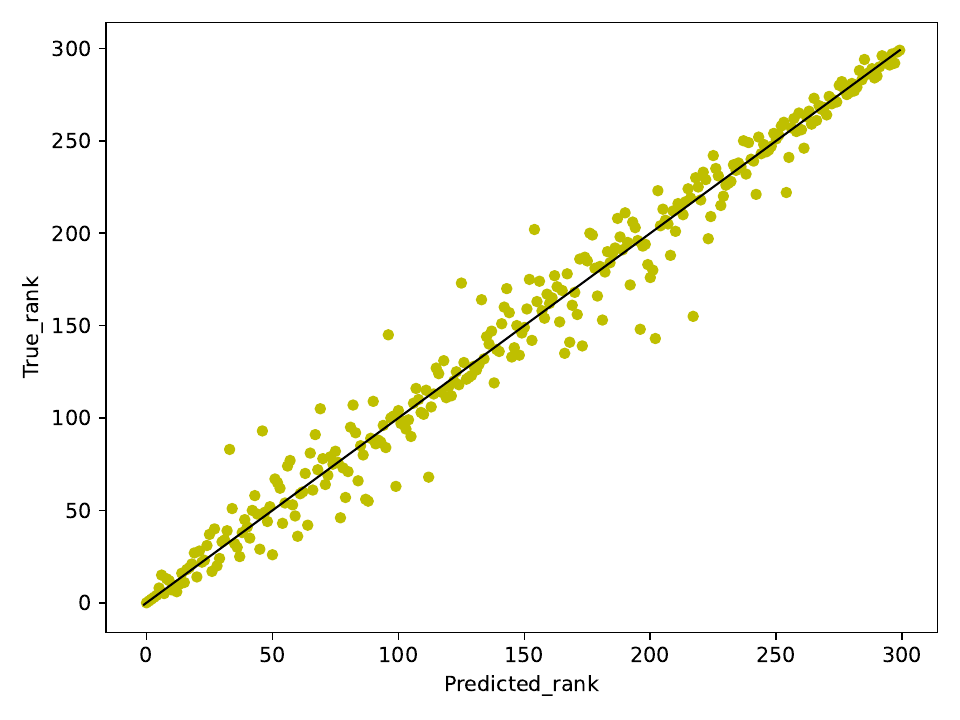}
\label{MLP}
}
\subfloat[KNN ($KTau$ = 0.88)]{
\includegraphics[width=0.22\linewidth]{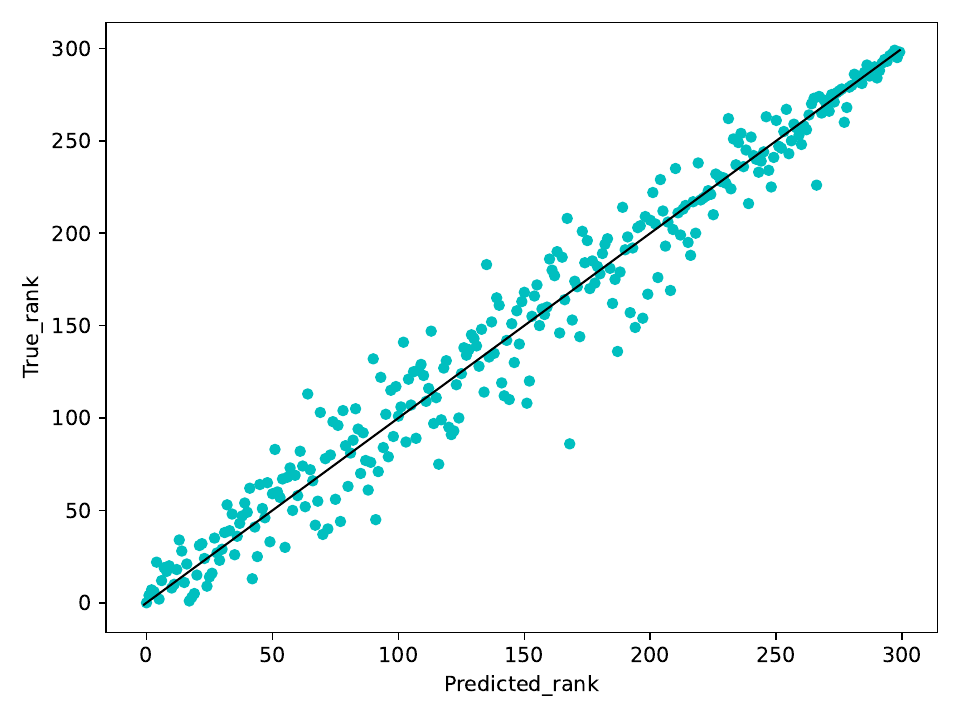}
\label{KNN}
}
\subfloat[RF ($KTau$ = 0.85)]{
\includegraphics[width=0.22\linewidth]{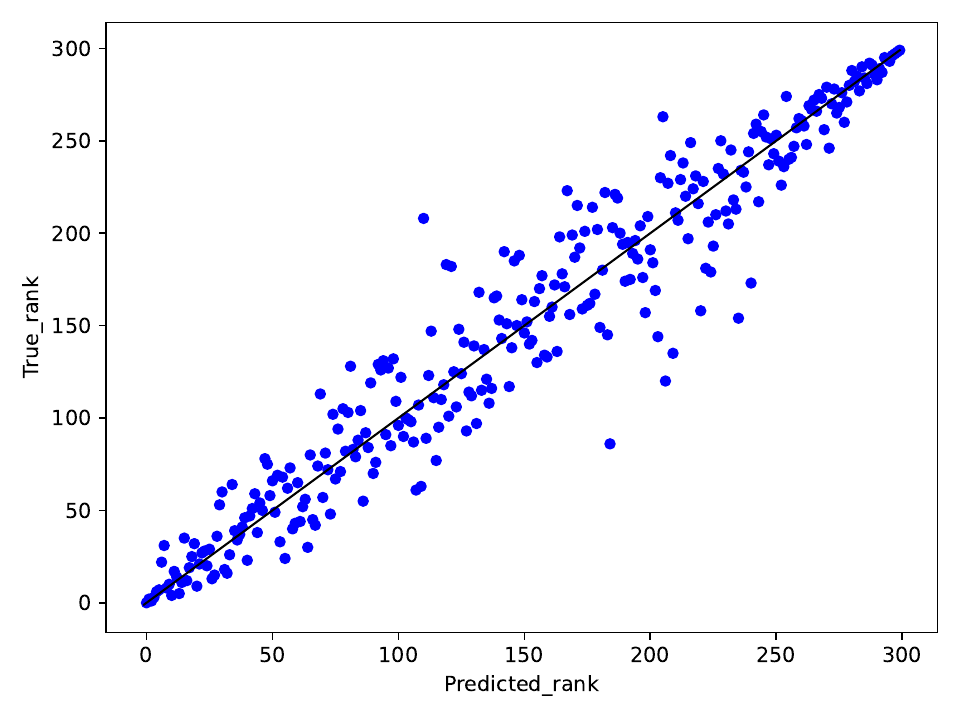}
\label{RF}
}
\caption{Search with different surrogate models. The top row shows the results of the search using four different models. The horizontal and vertical axes are MAdds and error, respectively. The bottom row shows the correlation between the predicted accuracy ranking and the true ranking using $KTau$.}
\label{fig:KTau}
\end{figure*}

\subsection{Performance of Multi-population Mechanism}\label{sec:resultsMP}
To further explore the advantages of different populations and design appropriate thresholds for the multi-population mechanism, we designed three experiments to determine the optimal use of the main and vice populations during evolution. We tested three configurations based on two populations: favoring the main population, favoring the secondary population, and using with equal probability. For each, we set a different probability for the two populations. When favoring the use of the main population, the probabilities of using the main and vice populations are 0.7 and 0.3. The values are reversed when favoring the vice population. In the last case, the probability of using both is 0.5. Fig. \ref{fig:threshold} shows the results of our experiments, and it can be seen that when the main population is used too much, the Pareto front converges better, but architectures are more concentrated; when the vice population is used too much, architectures are more uniformly distributed, but the convergence effect is poorer; and when the main and vice populations are used with equal probability, the search effect is somewhere balanced. Therefore, in order to achieve better search results, both convergence and diversity should be considered. In the early stage of search, both populations are used for the convergence and exploration in the search space. In the middle stage, the main population should be used to achieve fast convergence, and in the late stage, we should use the vice population for more diversity exploration.

The multi-population mechanism is an important strategy to increase the diversity of the solutions while ensuring convergence in this study. In order to verify its effectiveness, more experiments were conducted on the CIFAR-100 dataset in the \textbf{supplementary materials}. We separately conducted the architecture search with and without the multi-population mechanism, and the comparison between the two search processes is shown in Fig. \ref{fig:comparison}. The figures show the distributions of the architectures searched in the archive and new individuals generated during the evolutionary process. Fig. \ref{fig:comparison_wo} shows the search process without the multi-population mechanism and Fig. \ref{fig:comparison_MP} shows the search process using the multi-population mechanism.
Specifically, in both figures, the blue dots represent all the candidate architectures searched in the 5th iteration, and the red dots represent the new candidates generated till the 20th iteration. In addition, the blue and red lines represent the Pareto front of the 5th iteration and 20th iteration respectively.
In Fig. \ref{fig:comparison_wo}, it is obvious that except for the randomly initialized individuals, the new individuals generated in the later evolution are all almost distributed in the same solution region, and there is no obvious change in the Pareto front. However, in Fig. \ref{fig:comparison_MP}, we can see that the new individuals generated are uniformly distributed unlike without the multi-population mechanism. The Pareto front is also further updated in the excellent direction (lower error rate, smaller model size) and more promising individuals are explored. Evidently, the multi-population mechanism is better able to ensure the diversity of solutions and prevent them from falling into the local optima.

\subsection{Ablation Studies}\label{sec:surrogate_ablation}
The surrogate model is one of the critical components of SMEMNAS. Besides, the hyper-parameters in multi-population mechanism also affect the performance. We now present ablation studies examining different design choices for the surrogate model and other key hyper-parameters.
\textbf{Comparison of classification models:}
In order to construct more accurate surrogate models, we considered four classification models: SVM, random forest, k-nearest neighbors, and multi-layer perceptron. We performed four groups of experiments on the CIFAR-100 dataset. For more intuitive comparisons, we computed the correlation of 300 fully trained and evaluated architectures during the search process with each surrogate model. We choose the Kendall's Tau coefficient ($KTau$) to show the correlation between their true rankings and predicted rankings. Fig. \ref{fig:KTau} shows the searched results using different models and the corresponding $KTau$. Architectures from the archive were used to train surrogate models. After that, new $K$ architectures were retained in each generation and used for testing. The figure shows the results of the last generation. From the $KTau$ and the distribution of individuals in the search process, it can be seen that SVM is the best classifier with very limited training samples. The new individuals generated during the evolution when using KNN and RF as surrogate models are worse and non-uniformly distributed, which may be caused by the inaccurate prediction at the early stage. In addition, the performance of MLP is similar to SVM as shown in the figure. But, SVM can take less time for training and achieve better results than MLP with the limited samples. Therefore, we choose SVM as the surrogate model.

\textbf{Efficiency of the surrogate model:} According to our experimental setup (initial population with 100 individuals, evolving for 25 generations), approximately thousands of offspring would be generated during the whole search process. If all these architectures are truly evaluated, it may take more than 10 GPU days. By employing the proposed  surrogate model, the search time is significantly reduced.

Compared to traditional surrogate models, we conducted data augmentation, expanding from $n$ samples to $n(n-1)/2$, which greatly increases the training data. Increased training data can effectively improve the reliability of the surrogate model. \tableref{tab:surrogate} shows the performance comparison of the proposed surrogate model with the traditional regression surrogate models. We used 300 samples for training models, and in each experiment each surrogate model was trained with the same data. We trained three models with the proposed classification-based method (SVM, RF, and KNN), and we trained another three models with the traditional regression-based method (MLP, Decision Tree, and AdaBoost). 
Table \ref{tab:surrogate} highlights that while the training and prediction time of the proposed surrogate model surpasses that of traditional models, classification-based surrogate models demonstrate superior performance in terms of correlation coefficients. Among these, our SVM achieves the highest $KTau$ coefficient at 0.8113.
Therefore, the proposed surrogate model can provide more reliable prediction results and is more capable of retaining and choosing the excellent individuals during the search. Moreover, although it takes more time to train the proposed surrogate model, the training and prediction only take a few more minutes during the actual search, which is acceptable compared to the whole search time.

\begin{table}[t]
    \centering
    \caption{Comparison with different surrogate models. We record the Spearman coefficients ($\rho$), Kendall’s Tau coefficients ($KTau$) between the predicted accuracy rankings and the true accuracy rankings. The results are the average of ten experiments carried out.}
    \label{tab:surrogate}
    \scalebox{0.8}{
    \begin{tabular}{c|c|c|c|c|c}
    \hline\hline
        \textbf{Type} & \textbf{Model} & \textbf{\begin{tabular}[c]{@{}c@{}}Trainning\\time (s)\end{tabular}} & \textbf{\begin{tabular}[c]{@{}c@{}}Prediction\\time (s)\end{tabular}} & \textbf{$KTau$} & \textbf{$\rho$}  \\ \hline
          & \textbf{SVM (ours)} & 48.822 & 0.452 & \textbf{0.8113} & \textbf{0.9494} \\
         Classification & RF & 9.679 & 0.443& 0.7145 & 0.8731 \\
        & KNN & 62.763 & 0.425 & 0.6243 & 0.8161 \\ \hline
         & MLP & 58.785 & 0.0356 & 0.5417 & 0.6319 \\
         Regression & DecisionTree & 0.441 & 0.0396 & 0.5605 & 0.7590 \\
         & AdaBoost & 0.225 & 0.0347 & 0.6998 & 0.8753 \\
        \hline\hline
    \end{tabular}
    }
\end{table}

         

\begin{table}[t]
    \centering
 \caption{The hypervolume values are achieved with different size of initial samples.}
\begin{tabular}{c|c|c|c|c}
    \hline\hline
\textbf{Initial size} & 25 & 50 & 100 & 150  \\ \hline
\textbf{HV} & 0.8313 & 0.8339 & \textbf{0.8363} & 0.8360 \\
        \hline\hline
    \end{tabular}
\label{fig:HV}
\end{table}

\begin{table}[t]
    \centering
\caption{Comparison with different sizes of initial samples to train surrogate model.}
\begin{tabular}{c|c|c|c|c}
    \hline\hline
\textbf{Initial size} & 25 & 50 & 100 & 150  \\ \hline
\textbf{$KTau$} & 0.4465 & 0.5846 & 0.6955 & 0.7472 \\ \hline
\textbf{$\rho$} & 0.6154 & 0.7485 & 0.8575 & 0.9123 \\
        \hline\hline
    \end{tabular}
\label{tab:paraN}
\end{table}

\textbf{Initial population size:} The initial population is important for both the training of the surrogate model and the subsequent evolution. In order to achieve a better balance between search efficiency and search quality, we tested the appropriate value of the initial population size, $N$, by conducting the following small-scale experiments on the CIFAR-10. We set different sizes of the initial population, \ie$N=25$, $50$, $100$ and $150$. We recorded the effect of different population sizes on the hypervolume (HV) value and the comparison results are shown in Table. \ref{fig:HV}. From the table, it can be seen that HV achieves the highest value when the population size $N=100$. Although the HV obtained at $N=150$ and at $N=100$ are almost equivalent, 50\% more individuals need to be evaluated when $N=150$, which is more time consuming. In addition, we tested the prediction performance of the initial surrogate model as well. As seen in Table \ref{tab:paraN}, the reliability of the surrogate model is low if the initial population size is too small, which is another reason for the unfavorable search results at $N=25$ and $50$. Besides, if the surrogate model has low reliability, the search results are prone to be less stable. Therefore, after comprehensive consideration, we choose $N=100$ as the size of the initial population.

\begin{figure}[t]
\centering
\subfloat[Architectures distribution of static threshold]{
\includegraphics[width=0.45\linewidth]{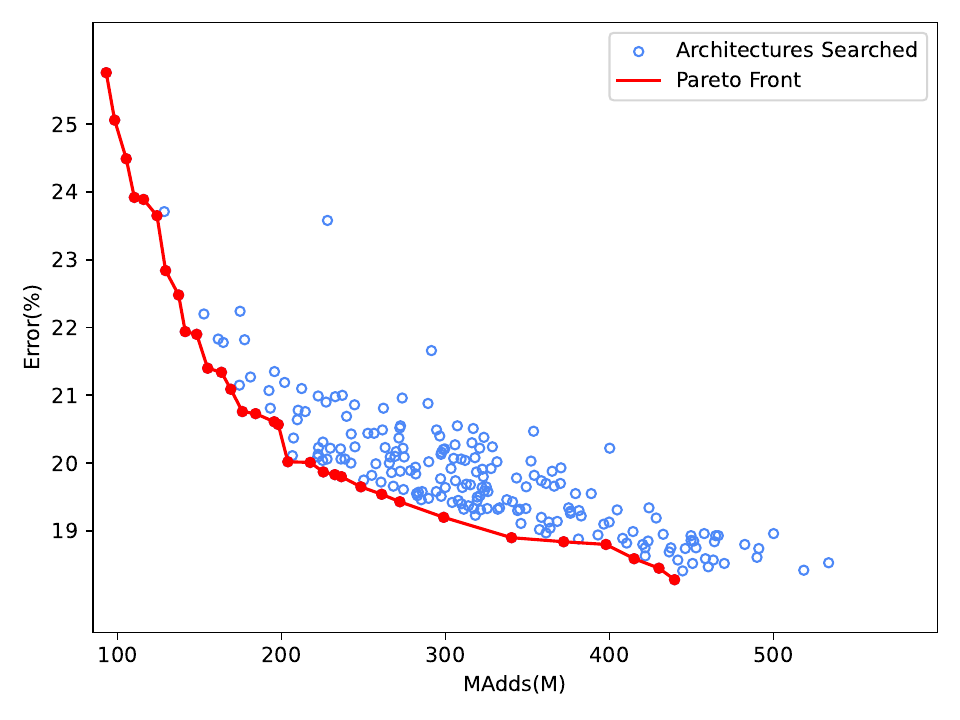}
\label{fig:static-1}
}
\hspace{2mm}
\subfloat[Architectures distribution of SMEMNAS]{
\includegraphics[width=0.45\linewidth]{figs/new/ParetoFront-8-6c.pdf}
\label{fig:static-2}
}
\caption{The comparison between static threshold and dynamic threshold on ImageNet. We plot the distribution of architectures in the final archive in two figures. The blue dots represent all the candidate architectures searched in the iterations, and the red line represent the Pareto front.}
\label{fig:static}
\end{figure}

\textbf{Dynamic threshold:} In SMEMNAS, the $Threshold$ that controls population generation is dynamically regulated. As can be observed from \myeqref{eq:Threshold} and \myeqref{eq:Threshold2}, when dynamic control is employed, even during phases that tend to favor diversity expansion, convergence is occasionally considered, and vice versa. To validate the effectiveness of this configuration, we designed a set of experiments with static control. We set the $Threshold$ value to consistently equal 0.3 in the early stage and 0.7 in the later stage, meaning that convergence is always considered in the early stage while diversity is always considered in the later stage. Fig. \ref{fig:static} demonstrates the results of the search Pareto front using static threshold compare with SMEMNAS. We can observe that the Pareto front of this configuration provides less diverse candidate solutions compared to the dynamic setting.
Further analysis of dynamic threshold can be found in the \textbf{supplementary materials} due to page limitations.

\section{Conclusion and Future Work}\label{sec:conclusion}

In this paper, we propose an evolutionary multi-objective algorithm that leverages a multi-population mechanism and a surrogate model based on pairwise comparison relations for neural architecture search. The proposed method utilizes a surrogate model to predict accuracy rankings rather than absolute accuracy values, effectively exploiting limited training samples and substantially reducing search overhead. Furthermore, a multi-population mechanism improves diversity in multi-objective NAS, increasing population diversity during evolution process, and accelerating algorithm convergence. Experiments on benchmark datasets verify the effectiveness of SMEMNAS. The final architectures discovered by SMEMNAS outperform most existing search methods.

In this work, our surrogate model only considers accuracy and ignores other network indicators. Single-objective surrogate models are not effectively adapted to multi-objective search frameworks. Future work will explore the multi-objective surrogate models to predict domination relations among architectures. Besides, the initial population significantly influences evolutionary algorithm performance. The proposed method simply uses random sampling to initialize the population, and the sampled architectures are non-uniformly distributed in the objective space, which may not provide an optimal foundation for subsequent evolutionary exploration. Therefore, future investigations will focus on efficient initialization methods, including diverse sampling strategies and Bayesian approaches. Additionally, to test model performance and applicability in different computer vision tasks, we will use a pretrained model on ImageNet as the backbone to test performance on other tasks such as object detection and semantic segmentation.
Furthermore, an important direction for future research is to systematically investigate the transferability of architectures discovered by SMEMNAS across significantly different domains and datasets. While our method demonstrates strong performance on image classification benchmarks (CIFAR-10, CIFAR-100, and ImageNet), understanding how well these architectures generalize to other visual recognition tasks, different data distributions, and even non-vision domains would provide valuable insights into our approach's robustness and generalizability. This cross-domain transferability analysis would help establish the practical applicability of SMEMNAS in diverse real-world scenarios.

\bibliography{cite}

@article{wang_idarts_2023,
	title = {{iDARTS}: {Improving} {DARTS} by node normalization and decorrelation discretization},
	volume = {34},
	number = {4},
	journal = {IEEE Transactions on Neural Networks and Learning Systems},
	author = {Wang, Huiqun and Yang, Ruijie and Huang, Di and Wang, Yunhong},
	year = {2023},
	pages = {1945--1957},
}

@article{yang_deeply_2025,
	title = {Deeply supervised block-wise neural architecture search},
	volume = {36},
	number = {2},
	journal = {IEEE Transactions on Neural Networks and Learning Systems},
	author = {Yang, An and Liu, Ying and Li, Chunguang and Ren, Qinyuan},
	year = {2025},
	pages = {2451--2464},
}

@article{tan_m2m-net_2024,
	title = {{M2M}-{Net}: multi-objective neural architecture search using dynamic {M2M} population decomposition},
	journal = {Neural Computing and Applications},
	author = {Tan, Zhiwen and Guo, Daqi and Chen, Junan and Chen, Lei},
	year = {2024},
}

@article{ma_defying_2025,
	title = {Defying multi-model forgetting in one-shot neural architecture search using orthogonal gradient learning},
	volume = {74},
	number = {5},
	journal = {IEEE Transactions on Computers},
	author = {Ma, Lianbo and Zhou, Yuee and Ma, Ye and Yu, Guo and Li, Qing and He, Qiang and Pei, Yan},
	year = {2025},
	pages = {1678--1689},
}

@article{garcia-garcia_continuous_2023,
	title = {Continuous cartesian genetic programming based representation for multi-objective neural architecture search},
	volume = {147},
	journal = {Applied Soft Computing},
	author = {Garcia-Garcia, Cosijopii and Morales-Reyes, Alicia and Escalante, Hugo Jair},
	year = {2023},
	pages = {110788},
}

@article{xue_surrogate_2025,
	title = {A surrogate model with multiple comparisons and semi-online learning for evolutionary neural architecture search},
	journal = {IEEE Transactions on Emerging Topics in Computational Intelligence},
	author = {Xue, Yu and Hu, Bohan and Neri, Ferrante},
	year = {2025},
	pages = {1--13},
}

@article{ding_bnas_2022,
	title = {{BNAS}: {Efficient} neural architecture search using broad scalable architecture},
	volume = {33},
	number = {9},
	journal = {IEEE Transactions on Neural Networks and Learning Systems},
	author = {Ding, Zixiang and Chen, Yaran and Li, Nannan and Zhao, Dongbin and Sun, Zhiquan and Chen, C. L. Philip},
	year = {2022},
	pages = {5004--5018},
}

@article{jiang_score_2025,
	title = {Score predictor-assisted evolutionary neural architecture search},
	doi = {10.1109/TETCI.2025.3526179},
	journal = {IEEE Transactions on Emerging Topics in Computational Intelligence},
	author = {Jiang, Pengcheng and Xue, Yu and Neri, Ferrante},
	year = {2025},
	pages = {1--15},
}

@article{zou_multiple_2024,
	title = {Multiple population alternate evolution neural architecture search},
	journal = {arXiv preprint arXiv:2403.07035},
	author = {Zou, Juan and Chu, Han and Xia, Yizhang and Xu, Junwen and Liu, Yuan and Hou, Zhanglu},
	year = {2024},
}

@inproceedings{liao_emt-nas_2023,
	title = {{EMT}-{NAS}: {Transferring} architectural knowledge between tasks from different datasets},
	booktitle = {Proceedings of the {IEEE}/{CVF} {Conference} on {Computer} {Vision} and {Pattern} {Recognition}},
	author = {Liao, Peng and Jin, Yaochu and Du, Wenli},
	year = {2023},
	pages = {3643--3653},
}

@article{song_multi-population_2024,
	title = {Multi-population evolutionary neural architecture search with stacked generalization},
	volume = {587},
	journal = {Neurocomputing},
	author = {Song, Changwei and Ma, Yongjie and Xu, Yang and Chen, Hong},
	year = {2024},
	pages = {127664},
}

@article{xu_co-evolutionary_2024,
	title = {A {Co}-evolutionary {Multi}-population {Evolutionary} {Algorithm} for {Dynamic} {Multiobjective} {Optimization}},
	volume = {89},
	journal = {Swarm and Evolutionary Computation},
	author = {Xu, Xin-Xin and Li, Jian-Yu and Liu, Xiao-Fang and Gong, Hui-Li and Ding, Xiang-Qian and Jeon, Sang-Woon and Zhan, Zhi-Hui},
	year = {2024},
	pages = {101648},
}

@article{lindauer_smac3_2022,
	title = {{SMAC3}: {A} versatile bayesian optimization package for hyperparameter optimization},
	volume = {23},
	number = {54},
	journal = {Journal of Machine Learning Research},
	author = {Lindauer, Marius and Eggensperger, Katharina and Feurer, Matthias and Biedenkapp, André and Deng, Difan and Benjamins, Carolin and Ruhkopf, Tim and Sass, René and Hutter, Frank},
	year = {2022},
	pages = {1--9},
}

@article{xue_gradient-guided_2024,
	title = {A gradient-guided evolutionary neural architecture search},
	journal = {IEEE Transactions on Neural Networks and Learning Systems},
	author = {Xue, Yu and Han, Xiaolong and Neri, Ferrante and Qin, Jiafeng and Pelusi, Danilo},
	year = {2024},
	pages = {1--13},
}

@article{xue_improved_2024,
	title = {Improved differentiable architecture search with multi-{Stage} progressive partial channel connections},
	volume = {8},
	number = {1},
	journal = {IEEE Transactions on Emerging Topics in Computational Intelligence},
	author = {Xue, Yu and Lu, Changchang and Neri, Ferrante and Qin, Jiafeng},
	year = {2024},
	pages = {32--43},
}

@article{xue_self-adaptive_2024,
	title = {Self-adaptive weight based on dual-attention for differentiable neural architecture search},
	volume = {20},
	number = {4},
	journal = {IEEE Transactions on Industrial Informatics},
	author = {Xue, Yu and Han, Xiaolong and Wang, Zehong},
	year = {2024},
	pages = {6394--6403},
}

@article{ma_pareto-wise_2024,
	title = {Pareto-wise ranking classifier for multi-objective evolutionary neural architecture search},
	volume = {28},
	number = {3},
	journal = {IEEE Transactions on Evolutionary Computation},
	author = {Ma, Lianbo and Li, Nan and Yu, Guo and Geng, Xiaoyu and Cheng, Shi and Wang, Xingwei and Huang, Min and Jin, Yaochu},
	year = {2024},
	pages = {570--581},
}

@article{tan_relativenas_2023,
	title = {{RelativeNAS}: {Relative} neural architecture search via slow-fast learning},
	volume = {34},
	number = {1},
	journal = {IEEE Transactions on Neural Networks and Learning Systems},
	author = {Tan, Hao and Cheng, Ran and Huang, Shihua and He, Cheng and Qiu, Changxiao and Yang, Fan and Luo, Ping},
	year = {2023},
	pages = {475--489},
}

@inproceedings{li_extensible_2023,
	title = {Extensible and efficient proxy for neural architecture search},
	booktitle = {Proceedings of the {IEEE}/{CVF} {International} {Conference} on {Computer} {Vision}},
	author = {Li, Yuhong and Li, Jiajie and Hao, Cong and Li, Pan and Xiong, Jinjun and Chen, Deming},
	year = {2023},
	pages = {6176--6187},
}

@article{gm_sequential_2024,
	title = {Sequential node search for faster neural architecture search},
	volume = {300},
	journal = {Knowledge-Based Systems},
	author = {G.m., Biju and Pillai, G. N.},
	year = {2024},
	pages = {112145},
}

@article{qiu_efficient_2023,
	title = {Efficient self-learning evolutionary neural architecture search},
	volume = {146},
	journal = {Applied Soft Computing},
	author = {Qiu, Zhengzhong and Bi, Wei and Xu, Dong and Guo, Hua and Ge, Hongwei and Liang, Yanchun and Lee, Heow Pueh and Wu, Chunguo},
	year = {2023},
	pages = {110671},
}

@article{yu_cyclic_2023,
	title = {Cyclic differentiable architecture search},
	volume = {45},
	number = {1},
	journal = {IEEE Transactions on Pattern Analysis and Machine Intelligence},
	author = {Yu, Hongyuan and Peng, Houwen and Huang, Yan and Fu, Jianlong and Du, Hao and Wang, Liang and Ling, Haibin},
	year = {2023},
	pages = {211--228},
}

@article{wei_npenas_2023,
	title = {{NPENAS}: {Neural} predictor guided evolution for neural architecture search},
	volume = {34},
	number = {11},
	journal = {IEEE Transactions on Neural Networks and Learning Systems},
	author = {Wei, Chen and Niu, Chuang and Tang, Yiping and Wang, Yue and Hu, Haihong and Liang, Jimin},
	year = {2023},
	pages = {8441--8455},
}

@article{zhang_gpu_2023,
	title = {{GPU} based genetic programming for faster feature extraction in binary image classification},
	journal = {IEEE Transactions on Evolutionary Computation},
	author = {Zhang, Rui and Sun, Yanan and Zhang, Mengjie},
	year = {2023},
	pages = {1--1},
}

@article{xie_architecture_2024,
	title = {Architecture augmentation for performance predictor via graph {Isomorphism}},
	volume = {54},
	number = {3},
	journal = {IEEE Transactions on Cybernetics},
	author = {Xie, Xiangning and Sun, Yanan and Liu, Yuqiao and Zhang, Mengjie and Tan, Kay Chen},
	year = {2024},
	pages = {1828--1840},
}

@article{lv_benchmarking_2023,
	title = {Benchmarking analysis of evolutionary neural architecture search},
	journal = {IEEE Transactions on Evolutionary Computation},
	author = {Lv, Zeqiong and Qian, Chao and Sun, Yanan},
	year = {2023},
	pages = {1--1},
}

@article{bi_network_2024,
	title = {Network attack prediction with hybrid temporal convolutional network and bidirectional {GRU}},
	volume = {11},
	number = {7},
	journal = {IEEE Internet of Things Journal},
	author = {Bi, Jing and Xu, Kangyuan and Yuan, Haitao and Zhang, Jia and Zhou, MengChu},
	year = {2024},
	pages = {12619--12630},
}

@article{tan_dynamic_2022,
	title = {Dynamic embedding projection-gated convolutional neural networks for text classification},
	volume = {33},
	number = {3},
	journal = {IEEE Transactions on Neural Networks and Learning Systems},
	author = {Tan, Zhipeng and Chen, Jing and Kang, Qi and Zhou, Mengchu and Abusorrah, Abdullah and Sedraoui, Khaled},
	year = {2022},
	pages = {973--982},
}

@inproceedings{zoph_neural_2017,
	title = {Neural architecture search with reinforcement learning},
	booktitle = {International {Conference} on {Learning} {Representations}},
	author = {Zoph, Barret and Le, Quoc V},
	year = {2017},
}

@inproceedings{zoph_learning_2018,
	title = {Learning transferable architectures for scalable image recognition},
	booktitle = {Proceedings of the {IEEE}/{CVF} {Conference} on {Computer} {Vision} and {Pattern} {Recognition}},
	author = {Zoph, Barret and Vasudevan, Vijay and Shlens, Jonathon and Le, Quoc V.},
	year = {2018},
	pages = {8697--8710},
}

@inproceedings{ghiasi_nas-fpn_2019,
	title = {{NAS}-{FPN}: {Learning} scalable feature pyramid architecture for object detection},
	booktitle = {Proceedings of the {IEEE}/{CVF} {Conference} on {Computer} {Vision} and {Pattern} {Recognition}},
	author = {Ghiasi, Golnaz and Lin, Tsung-Yi and Le, Quoc V.},
	year = {2019},
	pages = {7036--7045},
}

@inproceedings{nekrasov_fast_2019,
	title = {Fast neural architecture search of compact semantic segmentation models via auxiliary cells},
	booktitle = {Proceedings of the {IEEE}/{CVF} {Conference} on {Computer} {Vision} and {Pattern} {Recognition}},
	author = {Nekrasov, Vladimir and Chen, Hao and Shen, Chunhua and Reid, Ian},
	year = {2019},
	pages = {9126--9135},
}

@article{magalhaes_creating_2023,
	title = {Creating deep neural networks for text classification tasks using grammar genetic programming},
	volume = {135},
	journal = {Applied Soft Computing},
	author = {Magalhães, Dimmy and Lima, Ricardo H. R. and Pozo, Aurora},
	year = {2023},
	pages = {110009},
}

@inproceedings{tan_efficientnet_2019,
	title = {{EfficientNet}: {Rethinking} model scaling for convolutional neural networks},
	booktitle = {Proceedings of the 36th {International} {Conference} on {Machine} {Learning}},
	author = {Tan, Mingxing and Le, Quoc},
	year = {2019},
	pages = {6105--6114},
}

@inproceedings{pham_efficient_2018,
	title = {Efficient neural architecture search via parameters sharing},
	booktitle = {Proceedings of the 35th {International} {Conference} on {Machine} {Learning}},
	author = {Pham, Hieu and Guan, Melody and Zoph, Barret and Le, Quoc and Dean, Jeff},
	year = {2018},
	pages = {4095--4104},
}

@article{lu_multiobjective_2021,
	title = {Multiobjective evolutionary design of deep convolutional neural networks for image classification},
	volume = {25},
	number = {2},
	journal = {IEEE Transactions on Evolutionary Computation},
	author = {Lu, Zhichao and Whalen, Ian and Dhebar, Yashesh and Deb, Kalyanmoy and Goodman, Erik D. and Banzhaf, Wolfgang and Boddeti, Vishnu Naresh},
	year = {2021},
	pages = {277--291},
}

@inproceedings{lu_nsganetv2_2020,
	title = {{NSGANetV2}: {Evolutionary} multi-objective surrogate-assisted neural architecture search},
	booktitle = {Proceedings of the {European} {Conference} on {Computer} {Vision}},
	author = {Lu, Zhichao and Deb, Kalyanmoy and Goodman, Erik and Banzhaf, Wolfgang and Boddeti, Vishnu Naresh},
	year = {2020},
	pages = {35--51},
}

@article{sun_completely_2020,
	title = {Completely automated {CNN} architecture design based on blocks},
	volume = {31},
	doi = {10.1109/tnnls.2019.2919608},
	number = {4},
	journal = {IEEE Transactions on Neural Networks and Learning Systems},
	author = {Sun, Yanan and Xue, Bing and Zhang, Mengjie and Yen, Gary G.},
	year = {2020},
	pages = {1242--1254},
}

@inproceedings{baker_accelerating_2018,
	title = {Accelerating neural architecture search using performance prediction},
	booktitle = {International {Conference} on {Learning} {Representations}},
	author = {Baker, Bowen and Gupta, Otkrist and Raskar, Ramesh and Naik, Nikhil},
	year = {2018},
}

@article{wang_surrogate-assisted_2022,
	title = {Surrogate-assisted particle swarm optimization for evolving variable-length transferable blocks for image classification},
	volume = {33},
	doi = {10.1109/TNNLS.2021.3054400},
	number = {8},
	journal = {IEEE Transactions on Neural Networks and Learning Systems},
	author = {Wang, Bin and Xue, Bing and Zhang, Mengjie},
	year = {2022},
	pages = {3727--3740},
}

@inproceedings{lu_nsga-net_2019,
	title = {{NSGA}-{Net}: Neural architecture search using multi-objective genetic algorithm},
	booktitle = {Proceedings of the {Genetic} and {Evolutionary} {Computation} {Conference}},
	author = {Lu, Zhichao and Whalen, Ian and Boddeti, Vishnu and Dhebar, Yashesh and Deb, Kalyanmoy and Goodman, Erik and Banzhaf, Wolfgang},
	year = {2019},
	pages = {419--427},
}

@inproceedings{wang_evolving_2019,
	title = {Evolving deep neural networks by multi-objective particle swarm optimization for image classification},
	booktitle = {Proceedings of the {Genetic} and {Evolutionary} {Computation} {Conference}},
	author = {Wang, Bin and Sun, Yanan and Xue, Bing and Zhang, Mengjie},
	year = {2019},
	pages = {490--498},
}

@article{tong_neural_2022,
	title = {Neural architecture search via reference point based multi‐objective evolutionary algorithm},
	volume = {132},
	journal = {Pattern Recognition},
	author = {Tong, Lyuyang and Du, Bo},
	year = {2022},
	pages = {108962},
}

@inproceedings{liu_progressive_2018,
	title = {Progressive neural architecture search},
	booktitle = {Proceedings of the {European} {Conference} on {Computer} {Vision}},
	author = {Liu, Chenxi and Zoph, Barret and Neumann, Maxim and Shlens, Jonathon and Hua, Wei and Li, Li-Jia and Fei-Fei, Li and Yuille, Alan and Huang, Jonathan and Murphy, Kevin},
	year = {2018},
	pages = {19--34},
}

@inproceedings{cai_once_2020,
	title = {Once-{For}-{All}: {Train} one network and specialize it for efficient deployment},
	booktitle = {International {Conference} on {Learning} {Representations}},
	author = {Cai, Han and Gan, Chuang and Wang, Tianzhe and Zhang, Zhekai and Han, Song},
	year = {2020},
}

@inproceedings{luo_neural_2018,
	title = {Neural architecture optimization},
	volume = {31},
	booktitle = {Advances in {Neural} {Information} {Processing} {Systems}},
	author = {Luo, Renqian and Tian, Fei and Qin, Tao and Chen, Enhong and Liu, Tie-Yan},
	year = {2018},
}

@inproceedings{liu_darts_2018,
	title = {{DARTS}: {Differentiable} architecture search},
	booktitle = {International {Conference} on {Learning} {Representations}},
	author = {Liu, Hanxiao and Simonyan, Karen and Yang, Yiming},
	year = {2018},
}

@article{liu_survey_2022,
	title = {A survey on computationally efficient neural architecture search},
	volume = {1},
	number = {1},
	journal = {Journal of Automation and Intelligence},
	author = {Liu, Shiqing and Zhang, Haoyu and Jin, Yaochu},
	year = {2022},
	pages = {100002},
}

@article{sun_particle_2019,
	title = {A particle swarm optimization-based flexible convolutional autoencoder for image classification},
	volume = {30},
	number = {8},
	journal = {IEEE Transactions on Neural Networks and Learning Systems},
	author = {Sun, Yanan and Xue, Bing and Zhang, Mengjie and Yen, Gary G.},
	year = {2019},
	pages = {2295--2309},
}

@article{sun_surrogate-assisted_2020,
	title = {Surrogate-assisted evolutionary deep learning using an end-to-end random forest-based performance predictor},
	volume = {24},
	number = {2},
	journal = {IEEE Transactions on Evolutionary Computation},
	author = {Sun, Yanan and Wang, Handing and Xue, Bing and Jin, Yaochu and Yen, Gary G. and Zhang, Mengjie},
	year = {2020},
	pages = {350--364},
}

@inproceedings{howard_searching_2019,
	title = {Searching for {MobileNetV3}},
	booktitle = {Proceedings of the {IEEE}/{CVF} {International} {Conference} on {Computer} {Vision}},
	author = {Howard, Andrew and Sandler, Mark and Chen, Bo and Wang, Weijun and Chen, Liang-Chieh and Tan, Mingxing and Chu, Grace and Vasudevan, Vijay and Zhu, Yukun and Pang, Ruoming and Adam, Hartwig and Le, Quoc},
	year = {2019},
	pages = {1314--1324},
}

@inproceedings{sandler_mobilenetv2_2018,
	title = {{MobileNetV2}: {Inverted} residuals and linear bottlenecks},
	booktitle = {Proceedings of the {IEEE} {Conference} on {Computer} {Vision} and {Pattern} {Recognition}},
	author = {Sandler, Mark and Howard, Andrew and Zhu, Menglong and Zhmoginov, Andrey and Chen, Liang-Chieh},
	year = {2018},
	pages = {4510--4520},
}

@inproceedings{tan_mixconv_2019,
	title = {{MixConv}: {Mixed} depthwise convolutional kernels},
	booktitle = {British {Machine} {Vision} {Conference}},
	author = {Tan, Mingxing and Le, Quoc V.},
	year = {2019},
}

@inproceedings{chu_fairnas_2021,
	title = {{FairNAS}: {Rethinking} evaluation fairness of weight sharing neural architecture search},
	booktitle = {Proceedings of the {IEEE}/{CVF} {International} {Conference} on {Computer} {Vision}},
	author = {Chu, Xiangxiang and Zhang, Bo and Xu, Ruijun},
	year = {2021},
	pages = {12239--12248},
}

@inproceedings{yang_cars_2020,
	title = {{CARS}: {Continuous} evolution for efficient neural architecture search},
	booktitle = {Proceedings of the {IEEE}/{CVF} {Conference} on {Computer} {Vision} and {Pattern} {Recognition}},
	author = {Yang, Zhaohui and Wang, Yunhe and Chen, Xinghao and Shi, Boxin and Xu, Chao and Xu, Chunjing and Tian, Qi and Xu, Chang},
	year = {2020},
	pages = {1829--1838},
}

@article{xue_neural_2023,
	title = {Neural architecture search based on a multi-objective evolutionary algorithm with probability stack},
	volume = {27},
	number = {4},
	journal = {IEEE Transactions on Evolutionary Computation},
	author = {Xue, Yu and Chen, Chen and Słowik, Adam},
	year = {2023},
	pages = {778--786},
}

@inproceedings{guo_generalized_2022,
	title = {Generalized global ranking-aware neural architecture ranker for efficient image classifier search},
	booktitle = {Proceedings of the 30th {ACM} {International} {Conference} on {Multimedia}},
	author = {Guo, Bicheng and Chen, Tao and He, Shibo and Liu, Haoyu and Xu, Lilin and Ye, Peng and Chen, Jiming},
	year = {2022},
	pages = {3730--3741},
}

@article{ding_bnas-v2_2022,
	title = {{BNAS}-v2: {Memory}-efficient and performance-collapse-prevented broad neural architecture search},
	volume = {52},
	number = {10},
	journal = {IEEE Transactions on Systems, Man, and Cybernetics: Systems},
	author = {Ding, Zixiang and Chen, Yaran and Li, Nannan and Zhao, Dongbin},
	year = {2022},
	pages = {6259--6272},
}

@article{ding_stacked_2023,
	title = {Stacked {BNAS}: {Rethinking} broad convolutional neural network for neural architecture search},
	volume = {53},
	number = {9},
	journal = {IEEE Transactions on Systems, Man, and Cybernetics: Systems},
	author = {Ding, Zixiang and Chen, Yaran and Li, Nannan and Zhao, Dongbin and Chen, C. L. Philip},
	year = {2023},
	pages = {5679--5690},
}

@article{ming_two-stage_2022,
	title = {A two-stage evolutionary algorithm with balanced convergence and diversity for many-objective optimization},
	volume = {52},
	number = {10},
	journal = {IEEE Transactions on Systems, Man, and Cybernetics: Systems},
	author = {Ming, Fei and Gong, Wenyin and Wang, Ling},
	year = {2022},
	pages = {6222--6234},
}

@article{song_surrogate-assisted_2023,
	title = {A surrogate-assisted evolutionary framework with regions of interests-based data selection for expensive constrained optimization},
	volume = {53},
	number = {10},
	journal = {IEEE Transactions on Systems, Man, and Cybernetics: Systems},
	author = {Song, Zhenshou and Wang, Handing and Jin, Yaochu},
	year = {2023},
	pages = {6268--6280},
}

@article{yang_local-diversity_2023,
	title = {Local-diversity evaluation assignment dtrategy for decomposition-based multiobjective evolutionary algorithm},
	volume = {53},
	number = {3},
	journal = {IEEE Transactions on Systems, Man, and Cybernetics: Systems},
	author = {Yang, Shuling and Huang, Han and Luo, Fan and Xu, Yang and Hao, Zhifeng},
	year = {2023},
	pages = {1697--1709},
}

@inproceedings{peng_pre-nas_2022,
	title = {{PRE}-{NAS}: {Predictor}-assisted evolutionary neural architecture search},
	booktitle = {Proceedings of the {Genetic} and {Evolutionary} {Computation} {Conference}},
	author = {Peng, Yameng and Song, Andy and Ciesielski, Vic and Fayek, Haytham M. and Chang, Xiaojun},
	year = {2022},
	pages = {1066--1074},
}

@inproceedings{ye_-darts_2022,
	title = {$\beta$-{DARTS}: {Beta}-decay regularization for differentiable architecture search},
	booktitle = {Proceedings of the {IEEE}/{CVF} {Conference} on {Computer} {Vision} and {Pattern} {Recognition}},
	author = {Ye, Peng and Li, Baopu and Li, Yikang and Chen, Tao and Fan, Jiayuan and Ouyang, Wanli},
	year = {2022},
	pages = {10864--10873},
}

@inproceedings{xiao_shapley-nas_2022,
	title = {Shapley-{NAS}: {Discovering} operation contribution for neural architecture search},
	booktitle = {Proceedings of the {IEEE}/{CVF} {Conference} on {Computer} {Vision} and {Pattern} {Recognition}},
	author = {Xiao, Han and Wang, Ziwei and Zhu, Zheng and Zhou, Jie and Lu, Jiwen},
	year = {2022},
	pages = {11892--11901},
}

@inproceedings{chu_mixpath_2023,
	title = {{MixPath}: {A} unified approach for one-shot neural architecture search},
	booktitle = {Proceedings of the {IEEE}/{CVF} {International} {Conference} on {Computer} {Vision}},
	author = {Chu, Xiangxiang and Lu, Shun and Li, Xudong and Zhang, Bo},
	year = {2023},
	pages = {5972--5981},
}

@article{ding_nap_2022,
	title = {{NAP}: {Neural} architecture search with pruning},
	volume = {477},
	journal = {Neurocomputing},
	author = {Ding, Yadong and Wu, Yu and Huang, Chengyue and Tang, Siliang and Wu, Fei and Yang, Yi and Zhu, Wenwu and Zhuang, Yueting},
	year = {2022},
	pages = {85--95},
}

@inproceedings{tan_mnasnet_2019,
	title = {{MnasNet}: {Platform}-aware neural architecture search for mobile},
	booktitle = {Proceedings of the {IEEE}/{CVF} {Conference} on {Computer} {Vision} and {Pattern} {Recognition}},
	author = {Tan, Mingxing and Chen, Bo and Pang, Ruoming and Vasudevan, Vijay and Sandler, Mark and Howard, Andrew and Le, Quoc V.},
	year = {2019},
	pages = {2820--2828},
}

@inproceedings{ying_nas-bench-101_2019,
	title = {{NAS}-{Bench}-101: {Towards} reproducible neural architecture search},
	booktitle = {International {Conference} on {Machine} {Learning}},
	author = {Ying, Chris and Klein, Aaron and Christiansen, Eric and Real, Esteban and Murphy, Kevin and Hutter, Frank},
	year = {2019},
	pages = {7105--7114},
}

@article{tian_multistage_2021,
	title = {A multistage evolutionary algorithm for better diversity preservation in multiobjective optimization},
	volume = {51},
	number = {9},
	journal = {IEEE Transactions on Systems, Man, and Cybernetics: Systems},
	author = {Tian, Ye and He, Cheng and Cheng, Ran and Zhang, Xingyi},
	year = {2021},
	pages = {5880--5894},
}

@article{chen_diversity_2019,
	title = {A diversity ranking based evolutionary algorithm for multi-objective and many-objective optimization},
	volume = {48},
	journal = {Swarm and Evolutionary Computation},
	author = {Chen, Guoyu and Li, Junhua},
	year = {2019},
	pages = {274--287},
}

@article{huang_split-level_2023,
	title = {Split-level evolutionary neural architecture search with elite weight inheritance},
	journal = {IEEE Transactions on Neural Networks and Learning Systems},
	author = {Huang, Junhao and Xue, Bing and Sun, Yanan and Zhang, Mengjie and Yen, Gary G.},
	year = {2023},
	pages = {1--15},
}

@inproceedings{you2020greedynas,
  title={Greedynas: Towards fast one-shot nas with greedy supernet},
  author={You, Shan and Huang, Tao and Yang, Mingmin and Wang, Fei and Qian, Chen and Zhang, Changshui},
  booktitle={Proceedings of the IEEE/CVF Conference on Computer Vision and Pattern Recognition},
  pages={1999--2008},
  year={2020}
}

@inproceedings{guo2020single,
  title={Single path one-shot neural architecture search with uniform sampling},
  author={Guo, Zichao and Zhang, Xiangyu and Mu, Haoyuan and Heng, Wen and Liu, Zechun and Wei, Yichen and Sun, Jian},
  booktitle={Computer Vision--ECCV 2020: 16th European Conference, Glasgow, UK, August 23--28, 2020, Proceedings, Part XVI 16},
  pages={544--560},
  year={2020},
  organization={Springer}
}

@inproceedings{akiba_optuna_2019,
	address = {Anchorage AK USA},
	title = {Optuna: {A} next-generation hyperparameter optimization framework},
	doi = {10.1145/3292500.3330701},
	booktitle = {Proceedings of the 25th {ACM} {SIGKDD} {International} {Conference} on {Knowledge} {Discovery} \& {Data} {Mining}},
	author = {Akiba, Takuya and Sano, Shotaro and Yanase, Toshihiko and Ohta, Takeru and Koyama, Masanori},
	year = {2019},
	pages = {2623--2631},
}
\bibliographystyle{ieeetr}

\vspace{-30 pt}
\begin{IEEEbiography}
[{\includegraphics[width=1in,height=1.25in,clip,keepaspectratio]{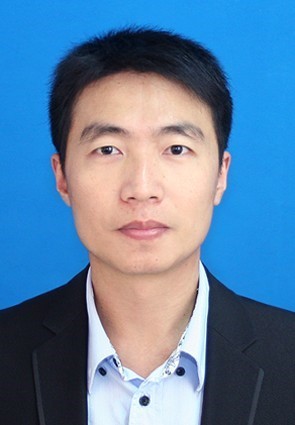}}]
{Yu Xue} (Senior Member, IEEE) received the Ph.D. degree from the School of Computer Science and Technology, Nanjing University of Aeronautics and Astronautics, Nanjing, China, in 2013. He was a Visiting Scholar with the School of Engineering and Computer Science, Victoria University of Wellington, Wellington, New Zealand, from August 2016 to August 2017. He was a Research Scholar with the Department of Computer Science and Engineering, Michigan State University, East Lansing, MI, USA, from October 2017 to November 2018. He is currently a Professor with the School of Software, Nanjing University of Information Science and Technology, Nanjing. His research interests include deep learning, evolutionary computation, machine learning, computer vision, and feature map selection.
\end{IEEEbiography}
\vspace{-30 pt}
\begin{IEEEbiography}
[{\includegraphics[width=1in,height=1.25in,clip,keepaspectratio]{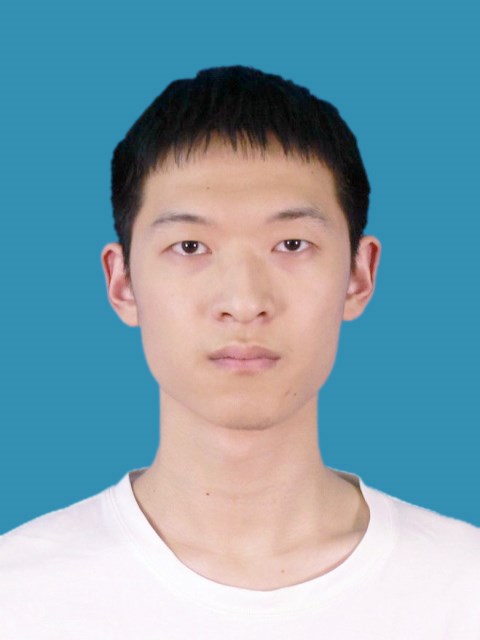}}]
{Pengcheng Jiang} (Graduate Student Member, IEEE) received the B.E. degree from Nanjing University of Information Science and Technology, China, in 2020. He is currently pursuing the Ph.D. degree with the School of Software in Nanjing University of Information Science and Technology, China. His current research interests include feature selection, evolutionary computation, neural architecture search, and model compression.
\end{IEEEbiography}
\vspace{-30 pt}
\begin{IEEEbiography}
[{\includegraphics[width=1in,height=1.25in,clip,keepaspectratio]{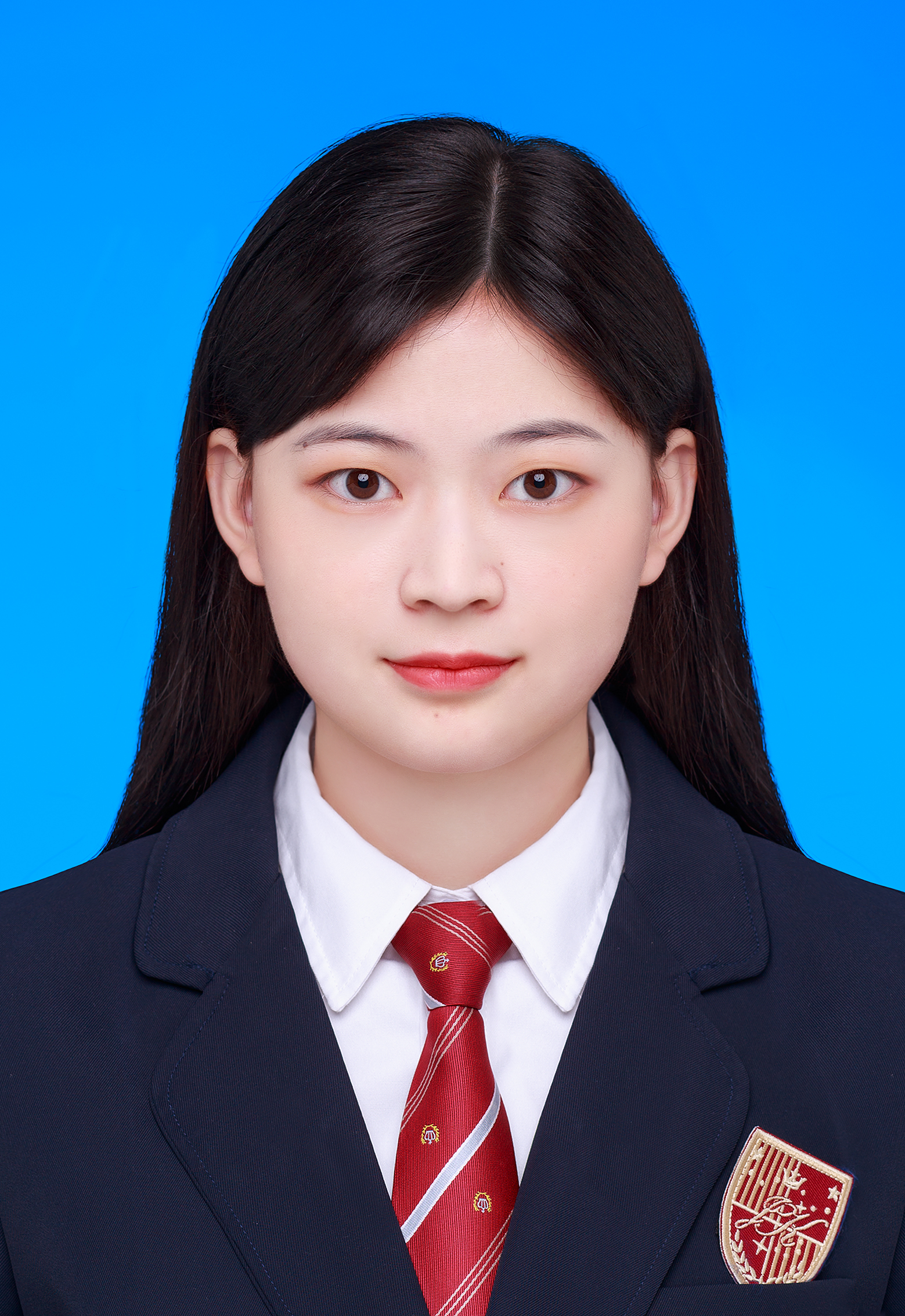}}]
{Chenchen Zhu} received the B.E. degree from Nanjing University of Information Science and Technology, China, in 2022, where she is currently pursuing a master’s degree. Her research interests include deep learning, multi-objective optimization and neural architecture search.
\end{IEEEbiography}
\vspace{-30 pt}
\begin{IEEEbiography}
[{\includegraphics[width=1in,height=1.25in,clip,keepaspectratio]{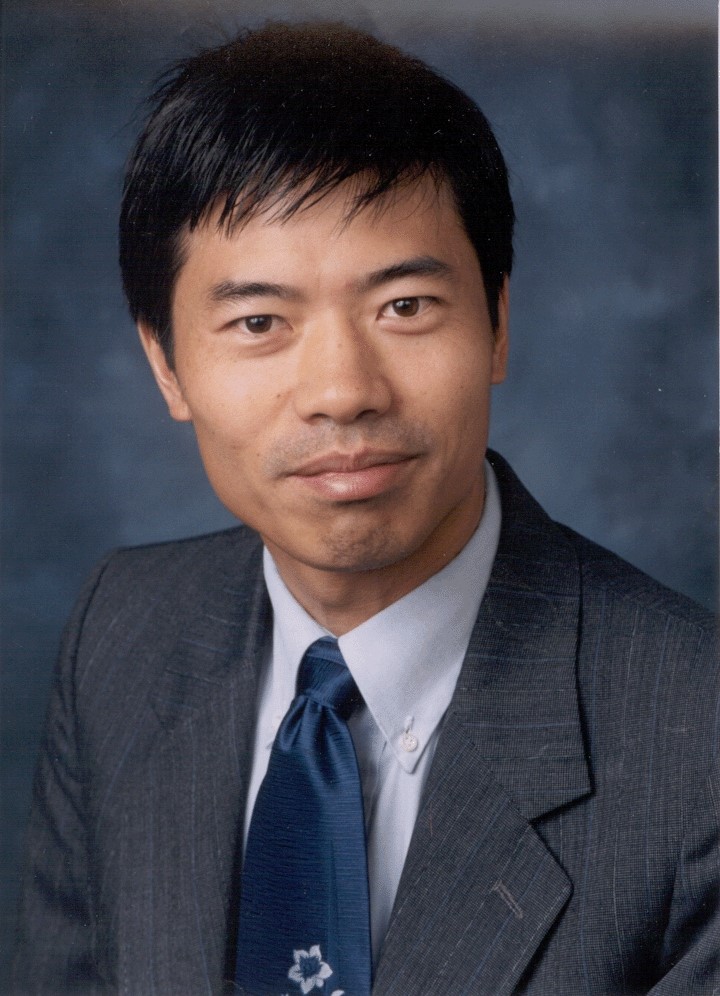}}] 
{MengChu Zhou} (Fellow, IEEE) received his Ph. D. degree from Rensselaer Polytechnic Institute, Troy, NY in 1990 and then joined New Jersey Institute of Technology where he has been Distinguished Professor since 2013. His interests are in Petri nets, automation, robotics, big data, Internet of Things, cloud/edge computing, and AI. He has 1200+ publications including 17 books, 850+ journal papers (650+ in IEEE transactions), 31 patents and 32 book-chapters. He is Fellow of IFAC, AAAS, CAA and NAI.
\end{IEEEbiography}
\vspace{-30 pt}
\begin{IEEEbiography}
[{\includegraphics[width=1in,height=1.25in,clip,keepaspectratio]{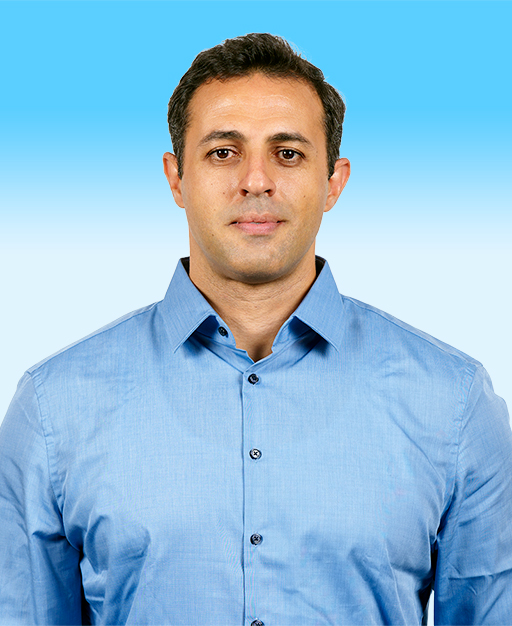}}] 
{Mohamed Wahib} is a team leader of the “High Performance Artificial Intelligence Systems Research Team” at RIKEN Center for Computational Science (R-CCS), Kobe, Japan. Prior to that he worked as is a senior scientist at AIST/TokyoTech Open Innovation Laboratory, Tokyo, Japan. He received his Ph.D. in Computer Science in 2012 from Hokkaido University, Japan. His research interests revolve around the central topic of highperformance programming systems, in the context of HPC and AI. He is actively working on several projects including high-level frameworks for programming traditional scientific applications, as well as high-performance AI.
\end{IEEEbiography}
\vspace{-30 pt}
\begin{IEEEbiography}
[{\includegraphics[width=1in,height=1.25in,clip,keepaspectratio]{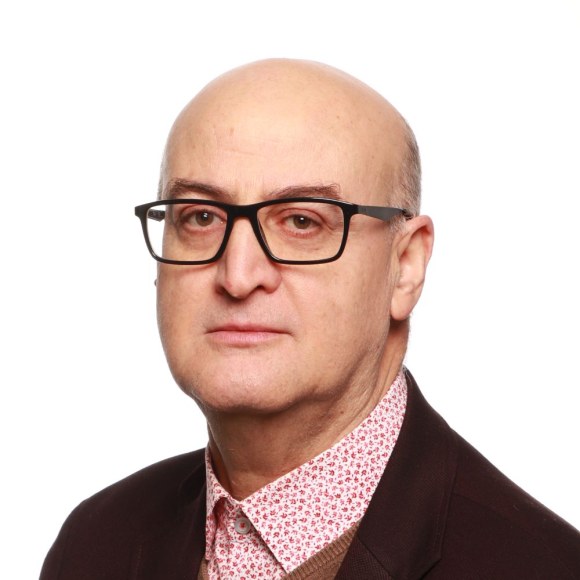}}] 
{Moncef Gabbouj} (Fellow, IEEE) was an Academy of Finland Professor. He is currently a Professor of information technology with the Department of Computing Sciences, Tampere University, Finland. His research interests include big data analytics, multimedia analysis, artificial intelligence, machine learning, pattern recognition, nonlinear signal processing, video processing, and coding. He is a fellow of Asia–Pacific Artificial Intelligence Association. He is a member of the Academia Europaea, the Finnish Academy of Science and Letters, and the Finnish Academy of Engineering Sciences. He is the Finland Site Director of the NSF IUCRC funded Center for Big Learning.
\end{IEEEbiography}

\includepdf[pages=-, fitpaper=true]{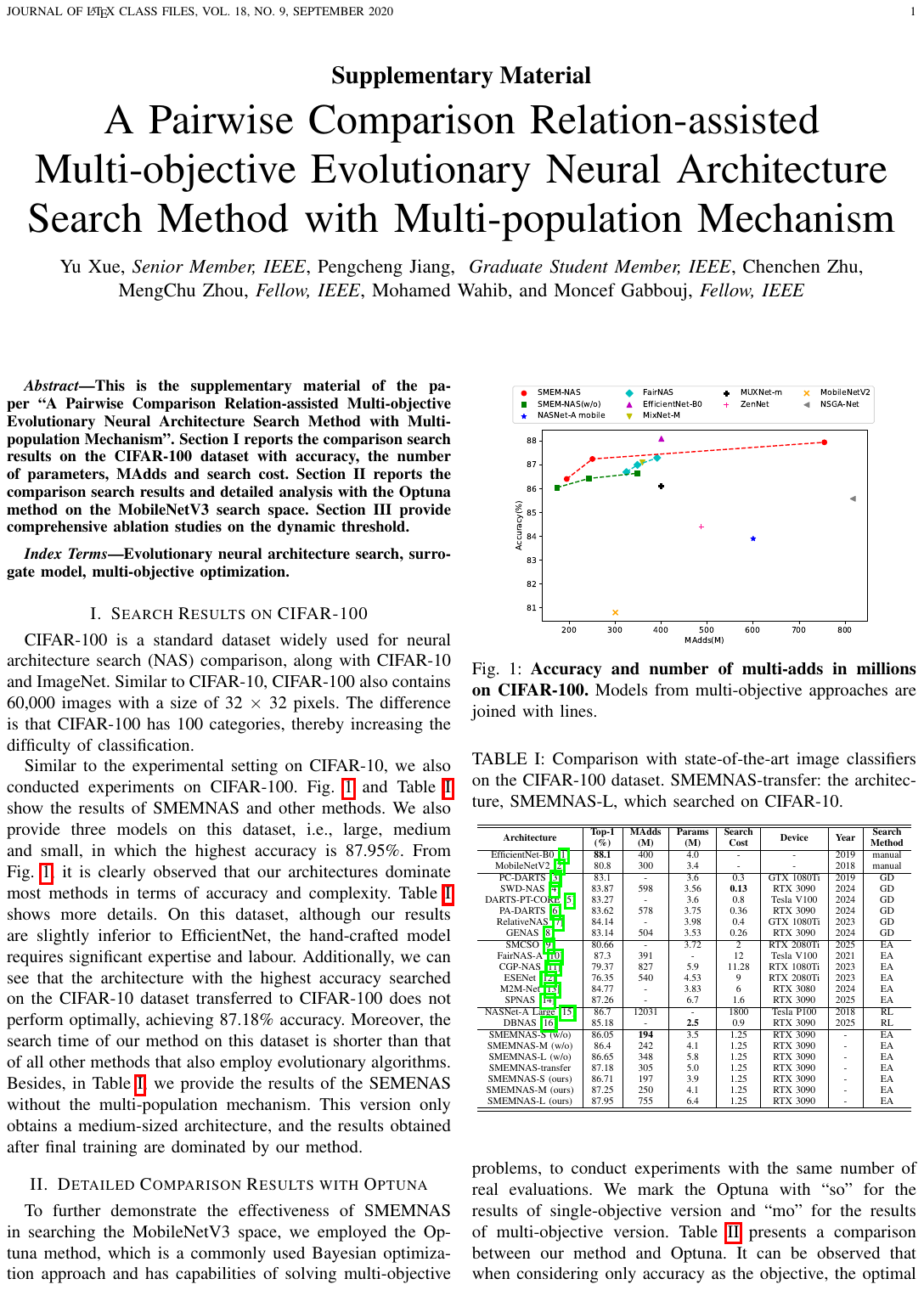}
\end{document}